\documentclass{article}

\usepackage{microtype}
\usepackage{graphicx}
\usepackage{subfigure}
\usepackage{booktabs} % for professional tables
\usepackage{multirow}
\usepackage{algorithm}
\usepackage{algorithmic}
\usepackage{csquotes}

\DeclareUnicodeCharacter{2212}{-}
\usepackage{amsmath,amsfonts,bm}

\def\eqref#1{equation~\ref{#1}}
\def\1{\bm{1}}

\DeclareMathAlphabet{\mathsfit}{\encodingdefault}{\sfdefault}{m}{sl}
\SetMathAlphabet{\mathsfit}{bold}{\encodingdefault}{\sfdefault}{bx}{n}

\newcommand{\E}{\mathbb{E}}

\newcommand{\R}{\mathbb{R}}

\usepackage{hyperref}

\usepackage[accepted]{icml2021}

\begin{document}

% keywords we want to have in the title:
% - multi-attribute
% - energy-based models
% - joint EBM
% - no factorized EBMs for me

\twocolumn[
% \icmltitle{Calibration, Uncertainty, and Conditional Synthesis for Multi-Attribute Datasets with EBMs}
% \icmltitle{Calibration, Uncertainty, and Conditional Synthesis for Multi-Attribute Data with Energy-Based Models}
% \icmltitle{Cut the Factorizations: Training Joint EBMs for Calibration, Uncertainty, and Conditional Synthesis}
% \icmltitle{Cut the Factorizing: Directly Training Joint Energy-Based Models for Calibration, Uncertainty, a
% \icmltitle{Directly Training Joint Energy-Based Models for Calibrated Prediction and Conditional Synthesis for Multi-Attribute Data}
\icmltitle{Directly Training Joint Energy-Based Models for Conditional Synthesis and Calibrated Prediction of Multi-Attribute Data}
% \icmltitle{Hybrid Discriminative-Conditional Synthesis for Multi-Attribute Data with Energy-Based Models}
% \icmltitle{No Factorizations for Me: Training Joint EBMs on Multi-Attribute Data with Calfibration, Uncertainty, and Conditional Synthesis}
% \icmltitle{Energy-Based Models for Calibration, Uncertainty, and Conditional Synthesis of Multi-Attribute Data}
% \icmltitle{Your Multi-Attribute Classifier is Secretly an Energy-Based Model}



\icmlsetsymbol{equal}{*}

\begin{icmlauthorlist}
% \icmlauthor{Aeiau Zzzz}{equal,to}
% \icmlauthor{Bauiu C.~Yyyy}{equal,to,goo}
% \icmlauthor{Cieua Vvvvv}{goo}
% \icmlauthor{Iaesut Saoeu}{ed}
% \icmlauthor{Fiuea Rrrr}{to}
% \icmlauthor{Tateu H.~Yasehe}{ed,to,goo}
% \icmlauthor{Aaoeu Iasoh}{goo}
% \icmlauthor{Buiui Eueu}{ed}
% \icmlauthor{Aeuia Zzzz}{ed}
% \icmlauthor{Bieea C.~Yyyy}{to,goo}
% \icmlauthor{Teoau Xxxx}{ed}
% \icmlauthor{Eee Pppp}{ed}
\icmlauthor{Jacob Kelly}{toronto_vector}
\icmlauthor{Richard Zemel}{toronto_vector}
\icmlauthor{Will Grathwohl}{toronto_vector}
% \icmlauthor{Anonymous}{anonymous}
\end{icmlauthorlist}

\icmlaffiliation{toronto_vector}{University of Toronto \& Vector Institute.
Code available at \href{https://github.com/jacobjinkelly/gibbs-jem}{\nolinkurl{github.com/jacobjinkelly/gibbs-jem}}}
% \icmlaffiliation{anonymous}{Anonymous}

% \icmlcorrespondingauthor{Anonymous}{anonymous}
\icmlcorrespondingauthor{Jacob Kelly}{jkelly@cs.toronto.edu}
% \icmlcorrespondingauthor{Will Grathwohl}{wgrathwohl@cs.toronto.edu}

\icmlkeywords{Machine Learning, ICML}

\vskip 0.3in
]



\printAffiliationsAndNotice{}  % leave blank if no need to mention equal contribution

\begin{abstract}

Multi-attribute classification generalizes classification, presenting new challenges for making accurate predictions and quantifying uncertainty.
We build upon recent work and show that architectures for multi-attribute prediction can be reinterpreted as energy-based models (EBMs).
While existing EBM approaches achieve strong discriminative performance, they are unable to generate samples conditioned on novel attribute combinations.
We propose a simple extension which expands the capabilities of EBMs to generating accurate conditional samples.
Our approach, combined with newly developed techniques in energy-based model training, allows us to directly maximize the likelihood of data and labels under the unnormalized joint distribution.
We evaluate our proposed approach on high-dimensional image data with high-dimensional binary attribute labels.
We find our models are capable of both accurate, calibrated predictions and high-quality conditional synthesis of novel attribute combinations.
\end{abstract}

\section{Introduction}

% narrow in on topic

Multi-attribute classification is a more general form of classification where each data example has a set of labels. 
% The task is to predict all of the attributes for a given example. 
% We can recover the classification problem by partitioning the set of attribute combinations and naming each attribute combination as a class.
% Multi-attribute labels may be a more generalizable form of supervision compared to single-attribute labels. 
% In many practical settings, it is often easier to collect multi-attribute labels than a single class-label. \JK{cite, expand upon}
% dig a hole
Models for multi-attribute prediction can be implemented similarly to models for single-attribute prediction, but face additional difficulties. 
Some attributes may be rare, causing issues related to severe class-imbalance. 
Attributes may be missing or only a subset may be observed for each example in the dataset. 
% Further, multi-attribute datasets may have many missing labels or only observe a subset of the attributes for each example. 
% For example, when dealing with many binary attributes, some attributes may be quite rare causing issues related to severe class-imbalance. 
In these settings, making calibrated predictions and quantifying uncertainty is especially important.

% fill the hole

Energy-Based Models (EBMs) present a flexible approach for representing uncertainty. 
% The ease with which structure may be incorporated and their relationship to physical systems has long made them popular in the broader scientific community~\citep{ingraham2019learning, du2020energy, noe2019boltzmann}. After a period of reduced interest from the machine learning community, EBMs have regained popularity as an approach for generative modeling~\citep{nijkamp2020anatomy,du2019implicit} and have found utility in many applications. 
Multi-attribute classifiers can be interpreted as Joint Energy-Based Models (JEM)~\citep{grathwohl2019your}. 
JEM re-purposes existing state-of-the-art classifier architectures to define an energy-based model of the joint distribution $p(x, y)$ of continuous images $x$ and discrete, 1-dimensional class labels $y$. 
JEM gives improved calibration, out-of-distribution detection, and adversarial robustness while retaining strong discriminative performance. 
\citet{grathwohl2019your} exploit the structure of the energy-function to marginalize out the label $y$ to uncover an EBM for $\log p(x)$ and a normalized $\log p(y|x)$ model. Then, the model can be trained to maximize the factorized likelihood ${\log p(x, y) = \log p(x) + \log p(y|x)}$. Techniques for training EBMs on continuous data are used to maximize the first term and the second term is optimized to minimize cross-entropy. 

% and has been extended to new modalities and application~\citep{he2021joint, hataya2021graph, grathwohl2020no,zhao2020joint}. 
% However, due to bias in MCMC-sampling~\citep{nijkamp2020learning, nijkamp2020anatomy}, training in this way presents challenges in conditional sampling. 
However, training JEM this way presents challenges in conditional sampling, especially if we want to condition on rare or novel attribute combinations. We propose a simple alternative approach to training JEM that directly maximizes the joint distribution $\log p(x, y)$. This requires sampling from the joint which can be challenging given the mixed continuous-discrete nature of the sampling problem.
% Notably though, solving this sampling task would enable the training of models simultaneously capable of discriminative modeling, conditional sampling, and semi-supervised learning. 
Leveraging recent improvements in EBM training~\citep{nijkamp2020anatomy, du2019implicit, du2020improved}, we find that we are able to directly train the joint model of data and labels.
% These models generate convincing conditional samples, and make accurate, calibrated predictions. 
Training in this way retains the benefits of joint modelling such as accurate discriminative performance and improved calibration, while also enabling these models to generate high-quality conditional samples of rare and novel attribute combinations. 

\section{Energy-Based Models}
An EBM parameterizes a probability distribution as
\begin{align}
    p_\theta(x) = \frac{e^{f_\theta(x)}}{Z(\theta)} \label{eq:ebm}
\end{align}
where $f_\theta: \mathbb{R}^D \rightarrow \mathbb{R}$ fully specifies the model and $Z(\theta) = \int e^{f_\theta(x)}\mathrm{d} x$ is the normalizing constant.
% which is not explicitly modeled.

While the flexibility of EBMs make them appealing, this flexibility comes with the cost of making sampling and likelihood evaluation difficult. EBMs are typically trained with gradient-based optimization using the following estimator for the gradient of the maximum likelihood objective:
\begin{align}
    \nabla_\theta \log p_\theta(x) = \nabla_\theta f_\theta(x) - \E_{p_\theta(x')}[\nabla_\theta f_\theta(x')].
    \label{eq:approxML}
\end{align}
Use of this estimator requires generating samples from $p_\theta(x)$. Since exact sampling is intractable, we resort to using approximate samples generated with MCMC~\citep{tieleman2008training}. Fortunately, a host of techniques have been developed to make training with MCMC efficient when using deep neural networks to define the energy-function~\citep{du2019implicit, du2020improved, nijkamp2019learning, nijkamp2020anatomy, xie2016theory}.

The preferred sampling approach for continuous data is Langevin Dynamics~\citep{welling2011bayesian} which updates samples with
\begin{align}
    x_{t+1} = x_t + \frac{\epsilon^2}{2 \lambda} \nabla_x f_\theta(x) + \epsilon \alpha, \qquad \alpha \sim N(0, I)
    \label{eq:sgld}
\end{align}
where the step-size $\epsilon$, and the temperature $\lambda$ are hyperparameters. Common values for image data are $\epsilon = 0.01$, $\lambda = \frac{1}{20,000}$. The low temperature is necessary to generate samples quickly enough for efficient training. 

\section{Joint Energy-Based Models}

We are given data in the form of pairs $(x, y)$ where $x \in \R^D$ and $y \in \{0, 1\}^K$. Multi-attribute prediction models parameterize the conditional distribution as ${p_\theta(y|x) = \prod_{k=1}^K p_\theta(y_k|x)}$. A parametric function ${f_\theta(x): \R^D \rightarrow \R^{K \times 2}}$ defines
\begin{align}
    p_\theta(y_k|x) &= \frac{\exp(f_\theta(x)[k][y_k])}{\exp(f_\theta(x)[k][0]) + \exp(f_\theta(x)[k][1])}    
    \label{eq:multi_label_cond}
\end{align}
where $f_\theta(x)[k][0], f_\theta(x)[k][1]$ are the logits for attribute $k$ taking values $0$ and $1$, respectively. Following \citet{grathwohl2019your}, we can use this same function $f_\theta$ to define an unnormalized model of the joint distribution over data points $x$ and attributes $y$:
\begin{align}
    p_\theta(x, y) &= 
    \frac{\prod_{i=1}^K\exp (f_\theta(x)[k][y_k])}{Z(\theta)}
    \label{eq:jem_joint}
\end{align}
where $Z(\theta)$ is the unknown normalizing constant. From this joint modeling interpretation, the form of $p_\theta(y|x)$ remains the same.

\subsection{Factored Energy-Based Models}

We can analytically marginalize out the labels $y$ in Equation \ref{eq:jem_joint} to obtain the unconditional distribution $p_\theta(x)$. Factoring the joint probability as $p_\theta(x, y) = p_\theta(x)p_\theta(y|x)$, we can train the first term with Equation \ref{eq:approxML} and the second term with standard cross-entropy~\citep{grathwohl2019your}. Ideally, training with this factorization or maximizing the joint log-likelihood directly would be identical, but since MCMC does not give exact samples, our gradient estimator is biased. This bias encourages the implicit distribution of the approximate MCMC sampler toward the data distribution~\citep{nijkamp2020learning}. Thus, the distribution to which we apply Equation \ref{eq:approxML} will impact the final model.

In particular, using the factorization from \citep{grathwohl2019your} impacts the final model by making the attribute-conditional distribution $p_\theta(x|y)$ hard to sample from directly. \citet{grathwohl2019your} circumvent this issue by relying on $p_\theta(x|y) \propto p_\theta(x)p_\theta(y|x)$ to generate attribute-conditional samples. While this approach works for low-dimensional $y$, it does not scale when $y$ is high-dimensional or highly structured. This is problematic if we wish to condition on a $y$ which is rare or, perhaps, does not appear in the training data, as we may never generate an $x$ where $y$ will be sampled from $p_\theta(y|x)$. %In these settings, the conditional distribution $p_\theta(y|x)$ often has limited support and it can be difficult to generate high-quality $x \sim p_\theta(x)$ with support in $p_\theta(y|x)$. 

\subsection{Directly Training the Joint Energy-Based Model}

To avoid these issues we propose to train by applying Equation \ref{eq:approxML} directly to the joint distribution $p_\theta(x, y)$ as
\begin{align*}
    \nabla_\theta \log p_\theta(x, y) = \nabla_\theta f_\theta(x, y) - \E_{p_\theta(x', y')}[\nabla_\theta f_\theta(x', y')]
\end{align*}
We find that training this way allows us to directly sample $x \sim p_\theta(x|y)$, allowing us to condition on rare and even novel attribute combinations. Of course, joint sampling from high-dimensional, unnormalized distributions is an incredibly difficult task but recent advances in gradient-based sampling and EBMs~\citep{welling2011bayesian, du2020energy, grathwohl2019your, grathwohl2021oops} have demonstrated that accurate samples can be generated from large-scale EBMs.  The main difficulty lies in generating samples from the joint distribution, which we describe in the following section.

\section{Training}
% In this section we outline our simple approach to sample from $p_\theta(x, y)$ where $x$ is continuous and $y$ is discrete. Defining efficient MCMC samplers for  distributions is still an open problem, but we found that our approach works well in practice.
When $x$ is held fixed, $p_\theta(y|x)$ is tractable and can be sampled from exactly.
When $y$ is held fixed, notice that
\begin{align}
    \label{eq:class_cond_dist}
    \log p_\theta(x|y) 
    &= \log p_\theta(x, y) - \log p_\theta(y)\nonumber \\
    &= f_\theta(x, y) - C(y)
\end{align}
where $C(y)$ is a constant that does not depend on $x$. Thus, $p_\theta(x|y)$ is an EBM defined on continuous $x$ given by evaluating $f_\theta(x, y)$ with $y$ fixed to the conditioning value. In this setting Langevin Dynamics has been successfully applied to generate samples.

We now introduce our approach to sample from $p_\theta(x, y)$ which we call Langevin-Within-Gibbs (LWG).
LWG works similarly to Gibbs sampling where we iteratively sample $y_{t+1} \sim p_\theta(y|x_t)$ and then $x_{t+1} \sim p_\theta(x|y_{t+1})$.
We can update our current sample $y_t$ exactly since $p_\theta(y|x_t)$ is tractable. 
However, the attribute-conditional distribution $p_\theta(x|y_{t+1})$ is unnormalized and intractable. Instead of sampling exactly from this conditional, we update our current sample $x_t$ using a Markov transition kernel applied to $p_\theta(x|y_{t+1})$, which we denote $\mathcal{T}_{c}(x_{t+1} | x_t, y_{t+1})$. In practice, this amounts to holding $y_{t+1}$ fixed and apply one step of Langevin dynamics to sample $x_{t+1}$.

It is likely that more involved approaches~\citep{zhou2019mixed} could lead to improved performance or more efficient sampling but we found this simple approach to work well for our applications. Pseudo-code for our proposed joint sampler can be seen in Algorithm \ref{alg:lwg}. We name our approach Gibbs-JEM, which involves sampling from JEM with LWG.
\vspace{-1.5em}
\begin{algorithm}[h]
  \caption{Joint Sampling} %\DD{Add step numbers?}}
  \label{alg:lwg}
\begin{algorithmic}
  \STATE {\bfseries Input:} EBM $p_\theta(x, y) \propto e^{f_\theta(x, y)}$, initial distribution $p_0(x)$, number of steps $T$
  \STATE {\bfseries Output:} Approximate samples $x_T, y_T \sim p_\theta(x, y)$
  \STATE Initialize samples $x_0 \sim p_0(x)$ \\
  $t = 0$
  \WHILE{$t < T$}
  \STATE Sample $y_{t+1} \sim p_\theta(y_{t+1} | x_t)$ \COMMENT{Categorical}
  \STATE Sample $x_{t+1} \sim \mathcal{T}_{c}(x_{t+1} | x_t, y_{t+1})$ \COMMENT{Langevin, Eq. \ref{eq:sgld}}\\
  $t = t + 1$
   \ENDWHILE
   \STATE Return $x_T, y_T$
\end{algorithmic}
\end{algorithm}
\vspace{-2em}
\section{Experiments}

We train models on $64\times64$ images of shoes from UT Zappos50K~\citep{utzappos2014, utzappos2017} and faces from CelebA~\citep{liu2015faceattributes}. 
We train a supervised baseline with cross-entropy, and a JEM baseline trained as in~\citet{grathwohl2019your}. Both baselines and Gibbs-JEM use the same architecture; the models only differ in how they are trained. Training details are available in Appendix \ref{appendix:experimental}.

\subsection{Prediction}

\begin{figure}[h]
    \centering
    \includegraphics[width=.23\textwidth, clip, 
    trim=2.2cm .1cm 2.4cm 1.2cm]
    {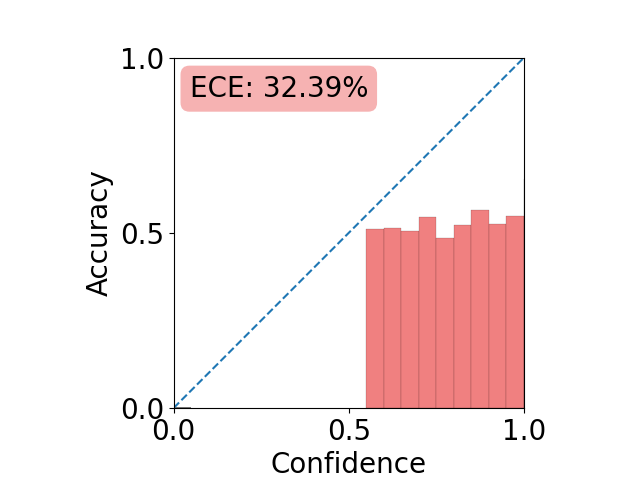}
     \includegraphics[width=.23\textwidth,
     clip, 
     trim=2.2cm .1cm 2.4cm 1.2cm]{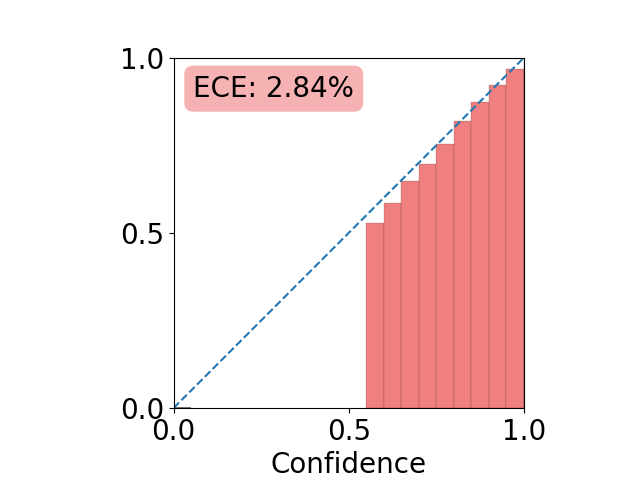}
    \vspace{-1em}
    \caption{Calibration (micro-averaged) on CelebA. Supervised (left) vs. Gibbs-JEM (right). ECE is Expected Calibration Error~\citep{guo17_ece}.}
    \label{fig:ece_main}
    % \vspace{-1.5em}
\end{figure}

\begin{figure}[h]
    \centering
    \includegraphics[width=.23\textwidth, clip, 
    trim=2.2cm .1cm 2.4cm 1.2cm]
    {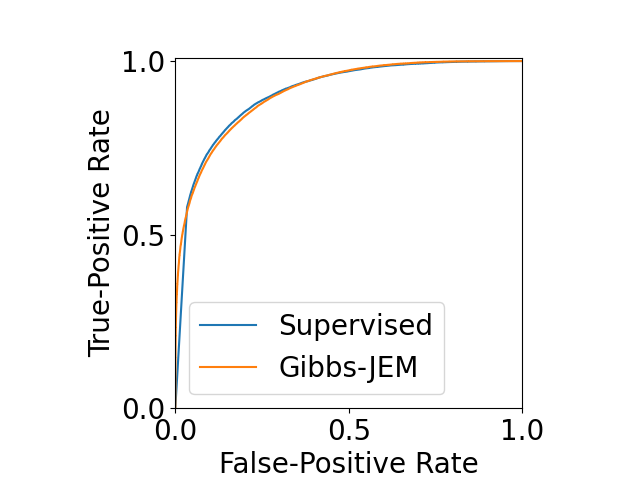}
    \includegraphics[width=.23\textwidth, clip, 
    trim=2.2cm .1cm 2.4cm 1.2cm]
    {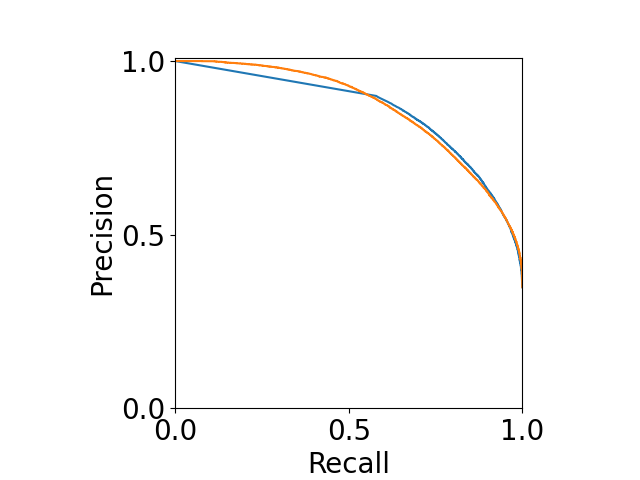}
    \vspace{-1em}
    \caption{Micro-averaged Receiver Operating Characteristic (left) and Precision-Recall (right) curves on CelebA.}
    \label{fig:roc_pr_main}
    % \vspace{-1cm}
\end{figure}

In Table \ref{table:results} we examine the predictive performance of our model. We find that JEM and Gibbs-JEM achieve competitive accuracy and F1 scores while giving notably more calibrated predictions, and superior AUROC and AUPRC.
We plot calibration diagrams and the Receiver Operating Characteristic and Precision-Recall curves of our model against the supervised baseline in Figures \ref{fig:ece_main} and \ref{fig:roc_pr_main} respectively. Additional evaluations and explanation of micro and macro averages are available in Appendix \ref{appendix:evaluation}.
\begin{table*}[h!]
    % \vspace{-1.5em}
    \centering
    \begin{tabular}{lcccccccccc}
         & \multicolumn{5}{c}{UTZappos} & \multicolumn{5}{c}{CelebA}  \\
         \cmidrule(lr){2-6}\cmidrule(lr){7-11} %\\
         & Accuracy & F1 & AUPRC & AUROC & ECE & Accuracy & F1 & AUPRC & AUROC & ECE \\
         \midrule
         Supervised 
         & \bf{91.87}  & \bf{82.74} & 88.19 & 95.97 & 26.39  
         & 84.53 & 77.07 & 82.77 & 90.89 & 32.39\\
         JEM
         & 90.05 & 78.68 & 86.73 & 94.32 & 7.428   
         & \bf{85.35} & \bf{77.29} & \bf{88.06} & \bf{92.27} & \bf{0.6987} \\
         Gibbs-JEM  
         & \bf{91.81} & 82.07 & \bf{91.53} & \bf{96.60} & \bf{0.2503}  
         & 84.14 & 74.64 & 86.39 & 91.17 & 2.838\\
         \bottomrule
    \end{tabular}
    \caption{Predictive performance of Gibbs-JEM versus baselines. AUPRC is computed using Average Precision. ECE is Expected Calibration Error. F1, AUPRC, AUROC, and ECE are micro-averaged over attributes; macro-averages have the same rank-order and are available in Appendix \ref{appendix:evaluation}.\label{table:results}}
\end{table*}
\subsection{Conditional Synthesis}
In addition to predictive performance, we find our models are also capable of high-quality conditional synthesis.
% We present results on conditional and unconditional image generation. 
We use a modified LWG sampler for conditional generation on a subset of the possible attributes, described in Appendix \ref{appendix:jem}. 

\begin{figure}[h]
    \centering
    \vspace{-.1cm}
    \includegraphics[width=.45\textwidth, clip, 
    trim=0.3cm 4cm 18.5cm 0cm]
    {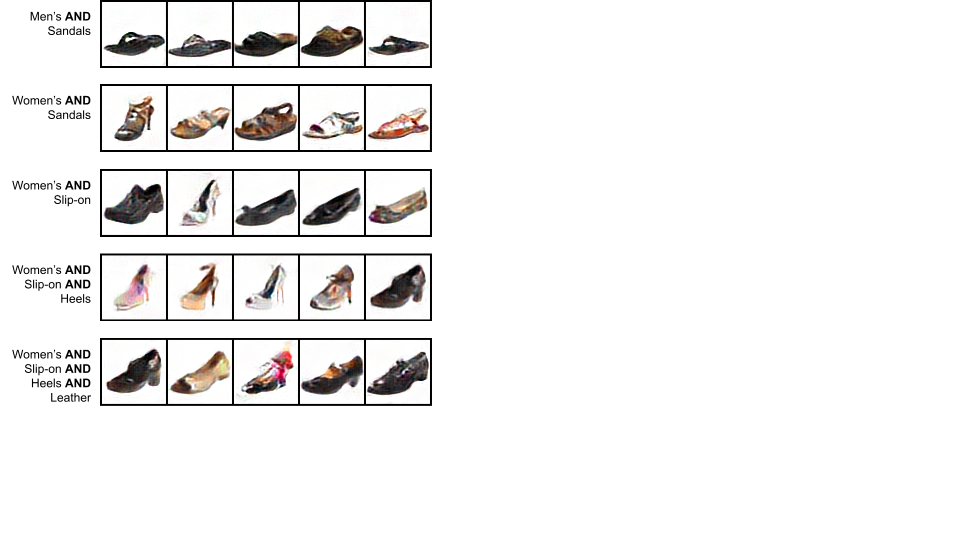}
    \vspace{-1em}
    \caption{Conditional samples on UT Zappos50K.\label{fig:utzappos_main}}
    \vspace{-1em}
\end{figure}

\begin{figure}[h]
    \centering
    \vspace{-1em}
    \includegraphics[width=.45\textwidth, clip, 
    trim=0.3cm 4cm 18.5cm 0cm]
    {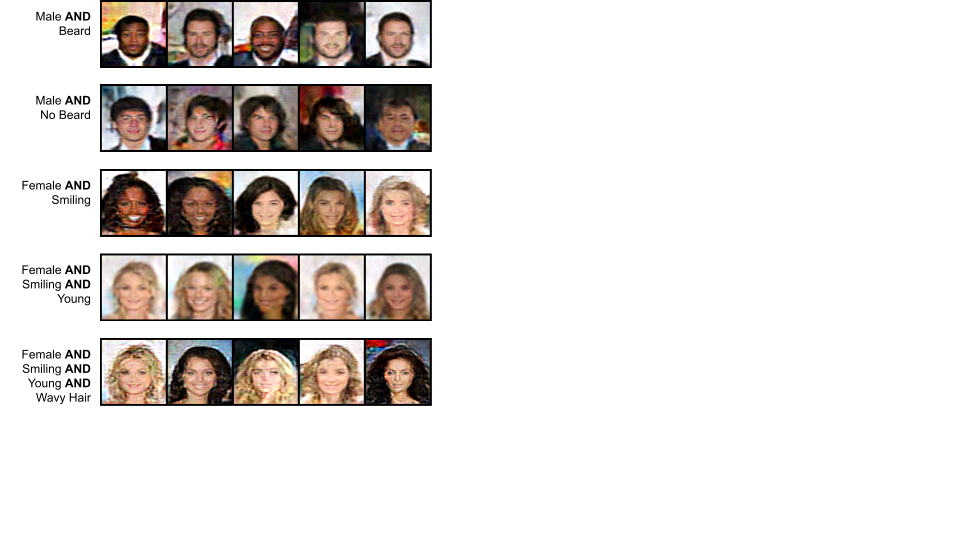}
    \vspace{-1em}
    \caption{Conditional samples on CelebA.\label{fig:celeba_main}}
    \vspace{-1em}
\end{figure}

\begin{figure}[h]
    \centering
    % \vspace{-1em}
    \includegraphics[width=.45\textwidth, clip, 
    trim=0.3cm 10.5cm 18.5cm 0cm]
    {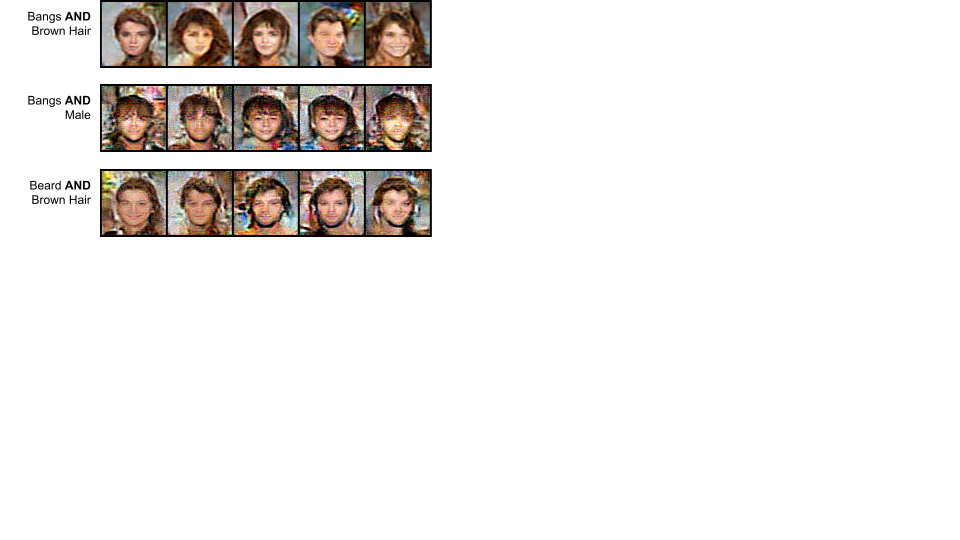}
    \vspace{-1em}
    \caption{Samples of novel attribute combinations on CelebA.\label{fig:celeba_zero_shot}}
    \vspace{-1em}
\end{figure}

% \begin{figure}[h]
%     \centering
%     \includegraphics[width=.45\textwidth, clip, 
%     trim=0cm 10cm 16cm 0cm]
%     {utzappos/single.png}
%     \includegraphics[width=.45\textwidth, clip, 
%     trim=0cm 10cm 16cm 0cm]
%     {utzappos/double.png}
%     \includegraphics[width=.45\textwidth, clip, 
%     trim=0cm 10cm 16cm 0cm]
%     {utzappos/successive.png}
%     \label{fig:utzappos_main}
%     \caption{Conditional samples on UTZappos.}
%     \vspace{-1em}
% \end{figure}

% \begin{figure}[h]
%     \centering
%     \vspace{-1em}
%     \includegraphics[width=.45\textwidth, clip, 
%     trim=0.3cm 13cm 18.8cm 0cm]
%     {celeba/double.png}
%     \includegraphics[width=.45\textwidth, clip, 
%     trim=0.3cm 10.6cm 18.8cm 0cm]
%     {celeba/successive.png}
%     \label{fig:celeba_main}
%     \vspace{-1em}
%     \caption{Conditional samples on CelebA.}
%     \vspace{-1em}
% \end{figure}

%\subsubsection{Image Quality}

In Figures \ref{fig:utzappos_main} and \ref{fig:celeba_main}, we demonstrate the quality of our model's samples when conditioning on successively more attributes. 
We compare the quality of these samples to those from the JEM baseline and examine the quality of other samples in more detail in Appendix \ref{appendix:samples}. 
% Additionally, we compare the quality of samples from the buffer compared to noise, using different conditional sampling methods, and more in Appendix \ref{appendix:samples}.

Next, we trained a Gibbs-JEM model on CelebA while holding out several attribute combinations. In Figure \ref{fig:celeba_zero_shot}, we plot samples from this model, conditioning on the held out attribute combinations. While sample quality has degraded, we find that the model is able to synthesize conditional samples of attributes it has seen separately, but never together, during training. A more thorough investigation of the model samples on held out attribute combinations is available in Appendix \ref{appendix:samples}.
% We also find the unconditional samples are unable to generate attribute combinations we desire (since there are too many possibilities).

%\subsubsection{Sample Diversity}

% We can generate diverse samples from the same noise using data augmentations and by conditioning on different logits.

% \subsubsection{Multiple Attributes}

% In Figure \ref{} we show that we can successively condition on more attributes for the same chain. While sample quality degrades slightly with the addition of new conditioning information, the new conditioning is reflected in the generated samples. 

% \subsubsection{Marginalization vs. Resampling}

% In Figure \ref{} we compare LWG sampling with to sampling by direct marginalization, we see that LWG and marginalizing the unobserved $y$ give similar results.

\section{Related Work}

Structured prediction problems have been a key application of EBMs. In this setting we wish to make predictions of highly structured $y$ given inputs $x$. To capture complex correlations, this is often phrased in an energy-minimization framework~\citep{gygli2017deep, belanger2017end} which has many similarities to the MCMC sampling we use. We believe our approach could be applied to these applications to add many of the benefits reported in \citet{grathwohl2019your}. 

Next are works that explore the unique capabilities of EBMs for challenging tasks such as continual learning~\citep{li2020energy} and compositional generation~\citep{DuComp2020}. We believe the techniques and architectures presented in this work could extend the range of problems these ideas can be applied to.

\section{Conclusion}
In this work we developed an approach to model the joint distribution of data and high-dimensional supervision using EBMs. We have demonstrated that our approach simultaneously achieves accurate and calibrated predictions and can perform high-quality conditional sampling, including of novel attribute combinations. Next steps include extending our approach to new data domains and settings with limited or partial supervision.

\section*{Acknowledgements}

Resources used in preparing this research were provided, in part, by the Province of Ontario, the Government of Canada through CIFAR, and companies sponsoring the Vector Institute.

\bibliography{ref}
\bibliographystyle{icml2021}

\appendix

\section{Joint Energy-Based Models}
\label{appendix:jem}

\subsection{Conditional Sampling}

After holding fixed the attributes we'd like to condition on, we are left with the problem of drawing joint samples of the data and the remaining \enquote{free} attributes. 
We refer to this as semi-conditional sampling. 
Semi-conditional sampling can be seen as a form of unconditional sampling once we've fixed the attributes to be conditioned on.
There are two approaches to unconditional sampling from our joint models. Thus there are two approaches to semi-conditional sampling.
The first approach involves drawing joint samples $x, y$ and then discarding the unneeded attributes $y$. We refer to this approach as \enquote{resampling}, since the attributes $y$ are being re-sampled. This form of sampling is how the model is trained. The other approach involves marginalizing out $y$ and drawing unconditional samples from $p_\theta(x)$. We refer this approach as \enquote{marginalizing} since the attributes $y$ are marginalized out. 
When applying these two approaches to unconditional sampling and semi-conditional sampling, the only difference lies in which attributes are \enquote{free}, as those are the ones which can be resampled or marginalized out.
% There are two approaches available for generating samples when conditioning on a subset of attributes. 
% The two approaches to semi-conditional sampling differ in how they handle the free attributes. 

In the following, let $c$ be the set of attribute indices we'd like to condition on, $\overline{c}$ its complement, $y_c$ the attribute vector $y$ with conditioned attributes $c$, and $y_{\overline{c}}$ the attribute vector $y$ with free attributes $\overline{c}$. For example, suppose we'd like to condition on attributes $1$ and $3$ set to $1$ and attribute $4$ set to $0$, and there are 5 attributes in total. Then $c = \{1, 3, 4\}, \overline{c}=\{2, 5\}$ and $y_c[1] = y_c[3] = 1$, $y_c[4] = 0$, and $y_c[i]$ is undefined for any $i \not\in c$, $y_{\overline{c}}$ is undefined for any $i \in c$.

\paragraph{Resampling} In Algorithm \ref{alg:modified_lwg}, we describe in more detail the procedure for semi-conditional sampling via resampling.

\begin{algorithm}[h]
  \caption{Semi-conditional Sampling (Resampling)} %\DD{Add step numbers?}}
  \label{alg:modified_lwg}
\begin{algorithmic}
  \STATE {\bfseries Input:} EBM $p_\theta(x, y) \propto e^{f_\theta(x, y)}$, initial distribution $p_0(x)$, number of steps $T$, conditioning information $y_c$
  \STATE {\bfseries Output:} Approximate samples $x_T, y_T \sim p_\theta(x, y_{\overline{c}}| y_c)$
  \STATE Initialize samples $x_0 \sim p_0(x)$ \\
  $t = 0$
  \WHILE{$t < T$}
  \FOR{$i \in c$}
    \STATE Copy $y_{t+1}[i] = y_c[i]$ \COMMENT{Conditioning attributes}
  \ENDFOR
  \FOR{$i \notin c$}
    \STATE Sample $y_{t+1}[i] \sim p_\theta(y_{t+1}[i] | x_t)$ \COMMENT{Free attributes}
  \ENDFOR
  
  \STATE Sample $x_{t+1} \sim \mathcal{T}_{c}(x_{t+1} | x_t, y_{t+1})$ \COMMENT{Langevin, Eq. \ref{eq:sgld}}\\
  $t = t + 1$
   \ENDWHILE
   \STATE Return $x_T, y_T$
\end{algorithmic}
\end{algorithm}

\paragraph{Marginalizing} Instead of resampling the attributes $y$ which aren't being conditioned on, we can instead marginalize them out. We sample from the following distribution:

\begin{align}
    \begin{split}
        p(x, y_{\overline{c}} | y_c) \propto
        &\bigg(\prod_{i \in c}\exp(f_\theta(x)[i][y_c[i]]\bigg) \cdot \\
        &\bigg(\prod_{i \not\in c}\exp(f_\theta(x)[0]) + \exp(f_\theta(x)[i][1])\bigg)
    \end{split}
    \label{eq:appendix_marginal_cond}
\end{align}

where the constant of proportionality is the normalizing constant. We write $\overline{\mathcal{T}}_{c}(x_{t+1} | x_t)$ for the Markov transition kernel which updates our current sample $x_t$ with one step of Langevin Dynamics according to the energy in Equation \ref{eq:appendix_marginal_cond}. We outline the sampling procedure for semi-conditional sampling via marginalizing in Algorithm \ref{alg:modified_lwg_marginal}.

\begin{algorithm}[h]
  \caption{Semi-conditional Sampling (Marginalizing)} %\DD{Add step numbers?}}
  \label{alg:modified_lwg_marginal}
\begin{algorithmic}
  \STATE {\bfseries Input:} EBM $p_\theta(x, y) \propto e^{f_\theta(x, y)}$, initial distribution $p_0(x)$, number of steps $T$, conditioning information $y_c$
  \STATE {\bfseries Output:} Approximate samples $x_T, y_T \sim p_\theta(x, y_{\overline{c}}| y_c)$
  \STATE Initialize samples $x_0 \sim p_0(x)$ \\
  $t = 0$
  \WHILE{$t < T$}
  \STATE Sample $x_{t+1} \sim \overline{\mathcal{T}}_{c}(x_{t+1} | x_t)$ \COMMENT{Langevin, Eq. \ref{eq:sgld}}\\
  $t = t + 1$
   \ENDWHILE
   \STATE Return $x_T, y_T$
\end{algorithmic}
\end{algorithm}

% \subsection{Maximum Likelihood Gradient}
% We give a short proof of Equation \ref{eq:marginal} adapted from~\citet{nijkamp2020learning}.
% Assuming we are only given $x^\prime$, then we can see
% \begin{align}
%     \nabla_\theta \log p_\theta(x^\prime)
%     &= \frac{\nabla_\theta p_\theta(x^\prime)}{p_\theta(x^\prime)} \nonumber \\
%     &= \frac{1}{p_\theta(x^\prime)}\nabla_\theta \int_\mathcal{Y} p_\theta(x^\prime, y) dy \tag{marginalize} \\
%     &= \frac{1}{p_\theta(x^\prime)}\int_\mathcal{Y} \nabla_\theta p_\theta(x^\prime, y) dy \tag{differentiate under the integral} \\
%     &= \frac{1}{p_\theta(x^\prime)}\int_\mathcal{Y} p_\theta(x^\prime, y) \nabla_\theta \log p_\theta(x^\prime, y) dy \tag{log-derivative trick} \\
%     &= \int_\mathcal{Y} \frac{p_\theta(x^\prime, y)}{p_\theta(x)} \nabla_\theta \log p_\theta(x^\prime, y) dy \nonumber \\
%     &= \int_\mathcal{Y} p_\theta(y | x^\prime) \nabla_\theta \log p_\theta(x^\prime, y) dy \nonumber \\
%     &= \E_{p_\theta(y|x^\prime)}\left[\nabla_\theta \log p_\theta(x^\prime, y)\right]  \nonumber \\
%     &= \E_{p_\theta(y|x^\prime)}\left[\nabla_\theta f_\theta(x^\prime, y)\right] - \E_{p_\theta(x, y)}\left[ \nabla_\theta f_\theta(x, y)\right]
% \end{align}

% where the last line is obtained by applying Equation \ref{eq:approxML} to the gradient of the joint log density $\nabla_\theta \log p_\theta(x^\prime, y)$.

\subsection{JEM for Multiple Attributes}

We introduce JEM~\citep{grathwohl2019your} for classifiers of multiple binary attributes. We used a softmax parameterization for ease of notation, but this can be extended to use a single logit for each binary attribute instead of the two that we use.

Following Equation \ref{eq:jem_joint}, we analytically marginalize out the binary attributes $y$ to obtain the unconditional distribution:

\begin{align*}
    p_\theta(x) &= \sum_{y_1,\dots, y_K} p(x, y_1, \dots, y_K) \\
    &\propto
    \sum_{y_1,\dots, y_K} \prod_{k=1}^K\exp f_\theta(x)[k][y_k] \\
    &\propto
    \sum_{y_1}\dots\sum_{y_K} \exp f_\theta(x)[1][y_1] \cdot \dots \cdot \exp f_\theta(x)[n][y_K] \\
    &\propto 
    \sum_{y_1}\exp f_\theta(x)[1][y_1] \cdot\dots\cdot \sum_{y_K}\exp f_\theta(x)[n][y_K] \\
    &\propto 
    \prod_{k=1}^K \sum_{y_k}\exp f_\theta(x)[k][y_k]
\end{align*}

where the constant of proportionality is $Z(\theta)$ from Equation \ref{eq:jem_joint}.

We can check that this joint interpretation gives the same parameterization of $p_\theta(y|x)$.

\begin{align*}
    p_\theta(y | x) 
    &= \frac{p_\theta(x, y)}{p_\theta(x)} \\
    &= \prod_{k=1}^K \frac{\exp f_\theta(x)[k][y_k]}{\sum_{y_k}\exp f_\theta(x)[k][y_k]} \\
    &=
    \prod_{k=1}^K p_\theta(y_k | x)
\end{align*}

This is the same as is given in Equation \ref{eq:multi_label_cond}, so we are done.

\section{Experimental Details}

\label{appendix:experimental}

\subsection{Training}

We use the Adam optimizer with default parameters $\beta_1 = 0.9, \beta_2 = 0.999$~\citep{kingma2014adam} throughout our experiments. We used a learning rate of $10^{-4}$ throughout our experiments. We found that the standard hyperparameters of $\epsilon=0.001$ and $\lambda=\frac{1}{20,000}$ worked well across datasets.

\paragraph{Sampling}

During training, we use Persistent Contrastive Divergence (PCD) and a replay buffer with a standard size of 10000 and replacement rate of $5\%$ throughout all of our experiments~\citep{du2019implicit}. At each sampling step, we clamp the sample to remain in the unit interval where the data lies, following \citet{du2020improved}.

\paragraph{Sampling Noise}
The initial distribution for Langevin dynamics is uniform with bounds equal to the per-dimension minimum and maximum values of one batch of data. 

\paragraph{Exponential Moving Average}

We keep an exponential moving average (EMA), as in \citet{du2020improved}, of the training parameters $\theta$ being optimized by applying the following update after each training iteration:
\begin{align*}
    \hat{\theta} = \mu\cdot \hat{\theta} + (1 - \mu)\cdot \theta.
\end{align*}
We initialize $\hat{\theta}_0 = \theta_0$ and set $\mu=0.999$ in our experiments. At test time we use the parameters $\hat{\theta}$.

\paragraph{Augmenting Sampling Chains}

Following \citet{du2020improved}, we apply data augmentations to buffer samples when sampling during training. We applied blurring when training on UT Zappos50k, and blurring and flipping transforms on CelebA. We use the same transforms with the same parameters and probability of application as in \citet{du2020improved}.

At test-time, for CelebA we apply augmentations every 60 steps. For UT Zappos50K we did not find data augmentations affected test-time sampling.

\paragraph{Missing Gradient Term in Contrastive Divergence}

We add the missing gradient term in contrastive divergence, as described in \citet{du2020improved}. We estimate this loss term following \citet{du2020improved} by backpropagating through the final step of Langevin dynamics. We weight the KL loss term with the standard weighting of $0.3$. We found that including this loss term greatly increases the diversity and quality of samples, especially when sampling from noise.

\paragraph{Reservoir Buffer}

Instead of the standard replay buffer, we use the reservoir buffer from \citet{du2020improved}. Samples from the reservoir buffer approximate uniform samples from the model over the course of its entire training, in contrast to the replay buffer, whose samples are biased towards more recent training iterations. We used the reservoir buffer when training on UT Zappos50K. We found that it increased the stability of sampling at the cost of slowing down learning.

\paragraph{Joint Persistent Contrastive Divergence}

Persistent Contrastive Divergence initializes sampling chains from a reservoir of past samples. Since we are training our models with joint samples, we likewise maintain a buffer of joint samples. This involves maintaining a parallel buffer which tracks the predicted attribute configurations of each corresponding sample in the buffer. We can then use this buffer of joint samples to conditionally initialize sampling chains at test-time to the desired attribute combination.

\paragraph{Number of Langevin Steps in Langevin-Within-Gibbs}

We found that increasing the number of Langevin steps used with Langevin-Within-Gibbs significantly increased the stability of joint sampling during training, at the cost of slightly decreasing discriminative performance. In particular, we used 2 steps instead of 1 step when training on CelebA, and reduced the number of Langevin-Within-Gibbs steps by half so as to keep the total number of Langevin steps constant. We used 40 Langevin steps for both UT Zappos50K and CelebA.

\paragraph{Initializing Langevin-Within-Gibbs}

We track the attributes in our replay buffer and initialize samples from buffer over the joint of images and labels. We also found it very important for stability to first sample the attributes conditioned on the data. Sampling the data conditioned on the random attributes we found to be significantly less stable.

\paragraph{Stability}

Training these models was still less stable than ideal. Regularizing the norm of the energies of data and generated samples has been found to be useful for improving stability and allowing for the use of larger step sizes past~\citep{du2019implicit, du2020improved}. In our experiments we were unable to regularized the norm of the energies of data and generated samples without severely impairing the performance of the model. We suspect that energy norm regularization prevents the discriminative performance of the model from improving. We were also unable to use large step sizes as used in \citet{du2020improved}, as this caused our models to diverge very fast, even when regularizing the energy norms. For this reason, we used a step size of 1 throughout our experiments. We believe that figuring out how to make the use of larger step sizes more stable is crucial for improving the quality of our model's samples.

\paragraph{Optimization}

We trained models and kept the weights which had the largest accuracy on a held-out set. For JEM and Gibbs-JEM models, we additionally checked that sampling had not diverged. Interestingly, we found that occasionally model accuracy would improve after sampling had diverged, and the model themselves would diverge shortly thereafter. We had limited success restarting Gibbs-JEM models with lower learning rates or increased number of steps after divergence. Interestingly, in contrast to results from \citet{grathwohl2019your}, we found JEM models to be quite stable in training. We suspect this is due to the new training techniques we applied that were developed in \citet{du2020improved}.

\paragraph{Architecture}

We use a variant of the convolutional architecture with 64 filters for 64x64 images from \citet{nijkamp2020learning}. We modified the final layer to be fully-connected and output logits according to the number of attributes. We replaced Leaky ReLU activations with the Swish activation~\citep{hendrycks2016gaussian, elfwing2018sigmoid, ramachandran2017swish}, which is named \verb|SiLU| in PyTorch. We used the following architecture for CelebA. The same architecture is used for UTZappos, except there are 38 outputs in the final layer (since UTZappos has 19 attributes, while CelebA has 23).

\begin{verbatim}
    (cnn): Sequential(
      (0): Conv2d(3, 64, 
            kernel_size=(3, 3), 
            stride=(1, 1), 
            padding=(1, 1))
      (1): SiLU(inplace=True)
      (2): Conv2d(64, 128, 
            kernel_size=(4, 4), 
            stride=(2, 2), 
            padding=(1, 1))
      (3): SiLU(inplace=True)
      (4): Conv2d(128, 256, 
            kernel_size=(4, 4), 
            stride=(2, 2), 
            padding=(1, 1))
      (5): SiLU(inplace=True)
      (6): Conv2d(256, 512, 
            kernel_size=(4, 4), 
            stride=(2, 2), 
            padding=(1, 1))
      (7): SiLU(inplace=True)
      (8): Conv2d(512, 512, 
            kernel_size=(4, 4), 
            stride=(2, 2), 
            padding=(1, 1))
      (9): SiLU(inplace=True)
      (10): Conv2d(512, 64, 
            kernel_size=(1, 1), 
            stride=(1, 1))
      (11): Flatten(start_dim=1, 
            end_dim=-1)
    )
    (mlp): Sequential(
      (0): Linear(in_features=1024, 
                out_features=128, 
                bias=True)
      (1): SiLU(inplace=True)
      (2): Linear(in_features=128, 
                out_features=128, 
                bias=True)
      (3): SiLU(inplace=True)
      (4): Linear(in_features=128, 
                out_features=46, 
                bias=True)
    )
\end{verbatim}

\paragraph{Compute}
We used PyTorch~\citep{NEURIPS2019_9015} and NumPy~\citep{harris2020array} throughout our experiments. We used NVIDIA T4 GPUs throughout our experiments.

\subsection{Data Processing}

We applied standard dequantization
\begin{align*}
    \tilde{x} = \frac{x * 255 + u}{256}, u \sim \mathcal{U}[0, 1]
\end{align*}
followed by adding Gaussian noise with standard deviation of $0.001$, and clamping to the interval $[0, 1]$ to both UT Zappos50K and CelebA.
We split 5000 random examples into a held out validation set for evaluation on both datasets.

On UT Zappos50K we filtered to attributes which had at least $10\%$ frequency of positivie examples in the dataset, while in CelebA we filtered to $13\%$. On UT Zappos50K we also dropped the attributes \enquote{HeelHeight}, \enquote{Insole}, and \enquote{ToeStyle} as they had too many missing attributes. After dropping these attributes, attributes \enquote{Gender}, \enquote{Material}, \enquote{Closure}, \enquote{Category}, and \enquote{SubCategory} had some missing values. We dropped any examples in the dataset which had missing values for these attributes. Many attributes in UT Zappos50K are categorical. We binarized these attributes by treating a $1-\text{of}-K$ one-hot categorical attribute as $K$ binary attributes.

For the held-out combinations experiment on CelebA, we held-out 11 paired-attribute combinations. This resulted in a held out set of 42,803 examples. The held out combinations were as follows:

\begin{itemize}
    \item Male AND Bangs
    \item Male AND Blond Hair
    \item Beard AND Bangs
    \item Beard AND Brown Hair
    \item Bangs AND Old
    \item Bangs AND Black Hair
    \item Bangs AND Brown Hair
    \item Bangs AND Blond Hair
    \item Old AND Black Hair
    \item Old AND Brown Hair
    \item Old AND Blond Hair
\end{itemize}

\section{Evaluation Metrics}
\label{appendix:evaluation}

\subsection{Averaging Across Attributes}

We used two approaches to average metrics across attributes. Macro-averaging computes the metric over each attribute separately, and then computes the average over attributes. Micro-averaging treats all attributes as the same attribute, and computes the metric as if there were $\verb|number of examples| \times \verb|number of attributes|$ examples. See \citet{lipton2014thresholding} for further discussion.

\subsection{Additional Evaluations}

We report macro-averaged metrics for both baselines and Gibbs-JEM in Table \ref{table:macro_results}.

\begin{table*}[h!]
    % \vspace{-1.5em}
    \centering
    \begin{tabular}{lcccccccc}
         & \multicolumn{4}{c}{UTZappos} & \multicolumn{4}{c}{CelebA}  \\
         \cmidrule(lr){2-5}\cmidrule(lr){6-9} %\\
         & F1 & AUPRC & AUROC & ECE & F1 & AUPRC & AUROC & ECE \\
         \midrule
         Supervised 
         & \bf{78.85} & 82.16 & 93.54 & 25.31
         &  \bf{69.93} & 72.35 & 86.73 & 32.03  \\
         JEM
         & 72.00 & 76.86 & 90.61 & 11.53   
         & 66.84 & \bf{75.64} & \bf{88.16} & \bf{1.421}\\
         Gibbs-JEM  
         & 75.84 & \bf{84.14} & \bf{94.09} & \bf{0.9106}  
         & 60.50 &  74.60 &  87.24 & 3.391 \\
         \bottomrule
    \end{tabular}
    \caption{Macro-averaged metrics.\label{table:macro_results}}
\end{table*}

\subsection{Additional Plots}

We show calibration diagrams, Receiver Operating Characteristic (ROC) curves and Precision-Recall (PR) curves for individual attributes on CelebA and UT Zappos50K, and micro-averagred on UT Zappos50K.

We found that the supervised baselines were notably less calibrated across attributes and datasets. We hypothesize that it is much easier for the supervised baseline to be overconfident in its predictions. We tuned the model performance by taking the weights with the best accuracy on the validation set. We found that supervised baselines needed to be trained for significantly more iterations to converge to the best validation accuracy. We also tried weighting the cross-entropy loss according to the frequencies of attributes in the training set, but this decreased accuracy.

We also find that the ROC and PR curves for supervised baselines are slightly misleading in some regions. Many predictions of the supervised baseline have confidence numerically 1, as is expected given the poor calibration of these models. Thus, in the PR curves, using a threshold very close to 1, for example $1 - 10^{-7}$, does not result in the model achieving a recall of 0. In these cases, the plots linearly interpolate the PR curve in the region between where recall is 0 and the recall where the largest threshold less than 1 is used. Since there are many predictions with confidence 1, this recall value is often very large (we've observed values greater than 0.5). Thus, where this occurs in some plots care should be taken in how they are interpreted.

\begin{figure}[h]
    \centering
    \includegraphics[width=.23\textwidth, clip, 
    trim=2.2cm .1cm 2.4cm 1.2cm]
    {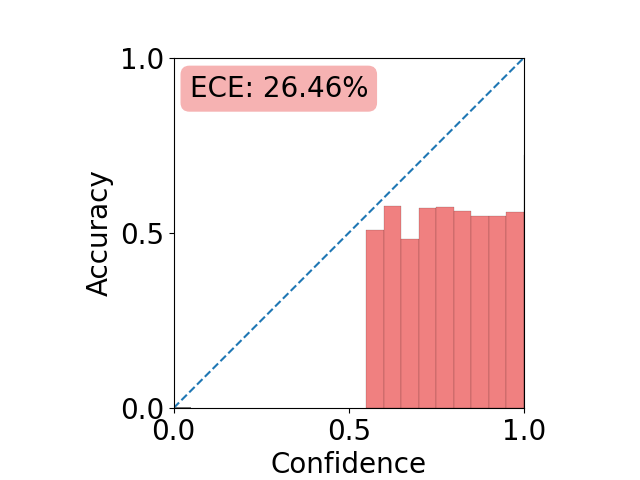}
     \includegraphics[width=.23\textwidth,
     clip, 
     trim=2.2cm .1cm 2.4cm 1.2cm]{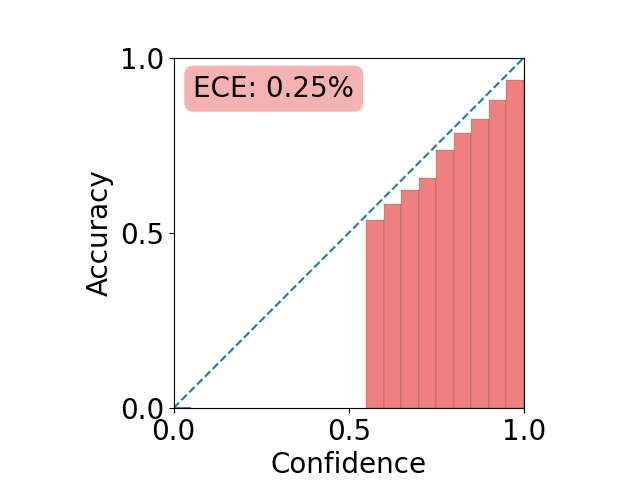}
    \vspace{-1em}
    \caption{Calibration (micro-averaged) on UT Zappos50K. Supervised (left) vs. Gibbs-JEM (right).}
\end{figure}

\begin{figure}[h]
    \centering
    \includegraphics[width=.23\textwidth, clip, 
    trim=2.2cm .1cm 2.4cm 1.2cm]
    {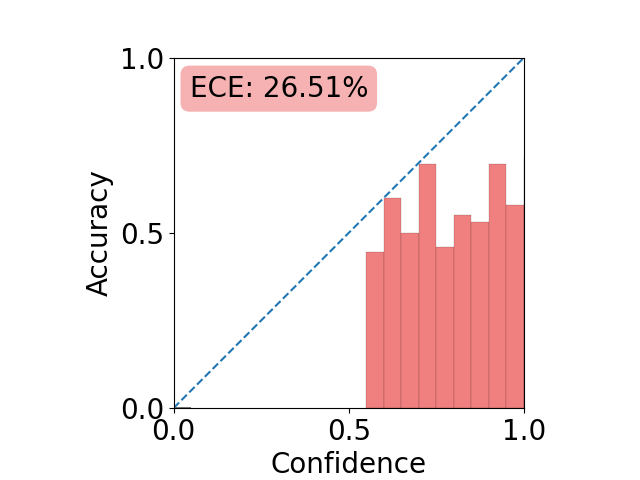}
     \includegraphics[width=.23\textwidth,
     clip, 
     trim=2.2cm .1cm 2.4cm 1.2cm]{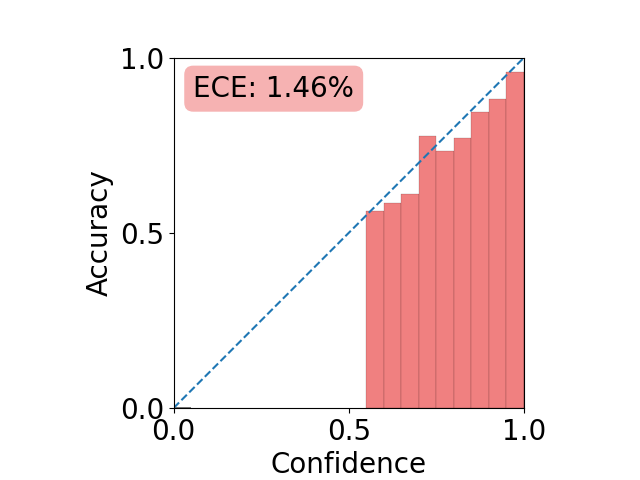}
    \vspace{-1em}
    \caption{Calibration on attribute \enquote{Men's} on UT Zappos50K. Supervised (left) vs. Gibbs-JEM (right).}
\end{figure}

\begin{figure}[h]
    \centering
    \includegraphics[width=.23\textwidth, clip, 
    trim=2.2cm .1cm 2.4cm 1.2cm]
    {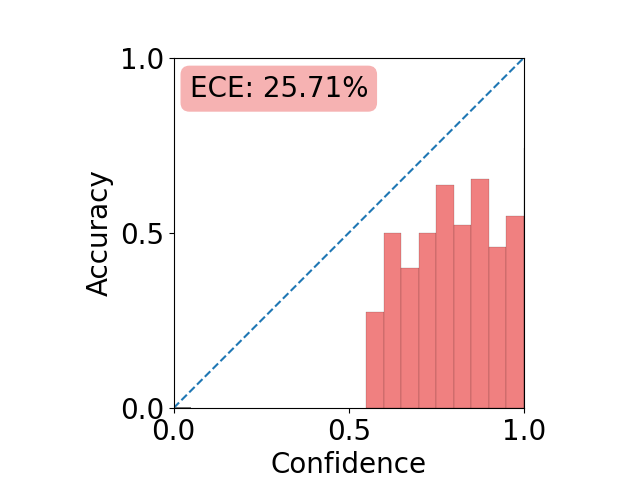}
     \includegraphics[width=.23\textwidth,
     clip, 
     trim=2.2cm .1cm 2.4cm 1.2cm]{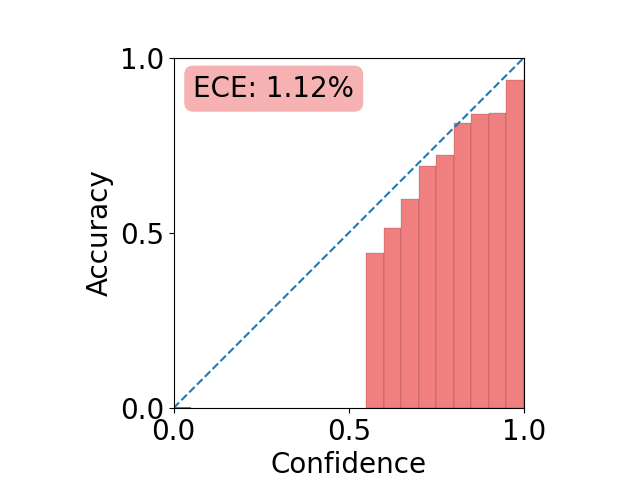}
    \vspace{-1em}
    \caption{Calibration on attribute \enquote{Girls'} on UT Zappos50K. Supervised (left) vs. Gibbs-JEM (right).}
\end{figure}

\begin{figure}[h]
    \centering
    \includegraphics[width=.23\textwidth, clip, 
    trim=2.2cm .1cm 2.4cm 1.2cm]
    {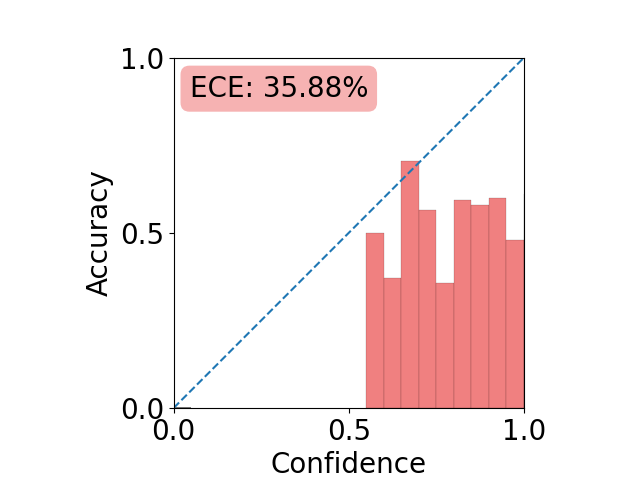}
     \includegraphics[width=.23\textwidth,
     clip, 
     trim=2.2cm .1cm 2.4cm 1.2cm]{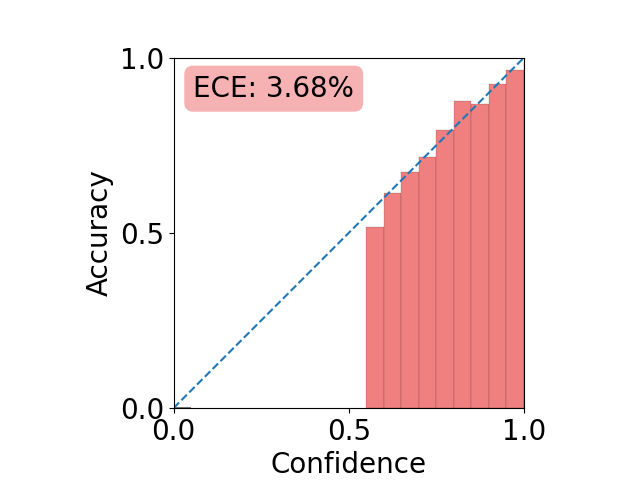}
    \vspace{-1em}
    \caption{Calibration on attribute \enquote{Arched eyebrows} on CelebA. Supervised (left) vs. Gibbs-JEM (right).}
\end{figure}

\begin{figure}[h]
    \centering
    \includegraphics[width=.23\textwidth, clip, 
    trim=2.2cm .1cm 2.4cm 1.2cm]
    {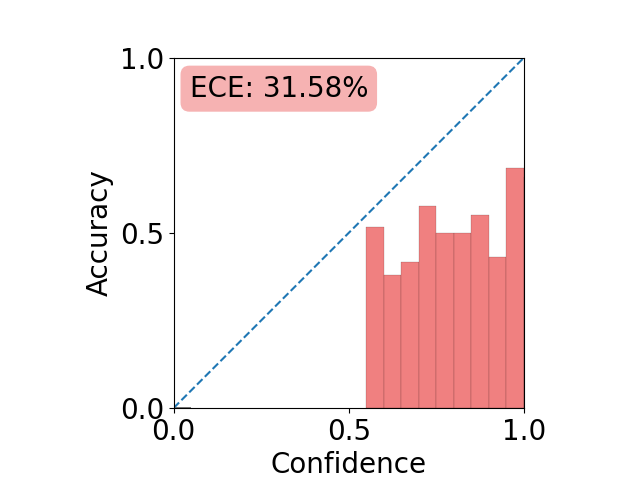}
     \includegraphics[width=.23\textwidth,
     clip, 
     trim=2.2cm .1cm 2.4cm 1.2cm]{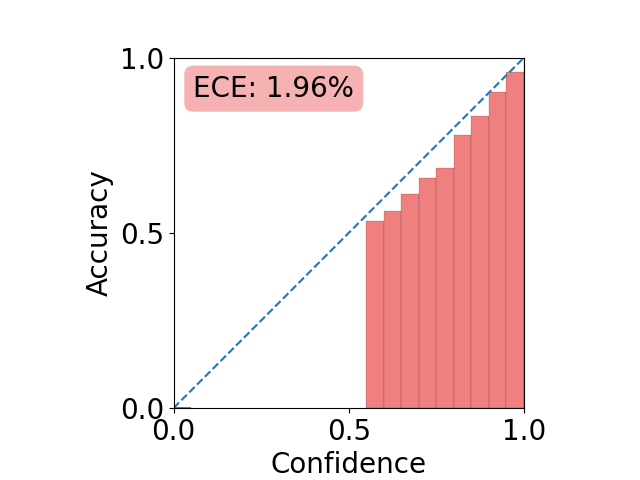}
    \vspace{-1em}
    \caption{Calibration on attribute \enquote{Bags under eyes} on CelebA. Supervised (left) vs. Gibbs-JEM (right).}
\end{figure}

\begin{figure}[h]
    \centering
    \includegraphics[width=.23\textwidth, clip, 
    trim=2.2cm .1cm 2.4cm 1.2cm]
    {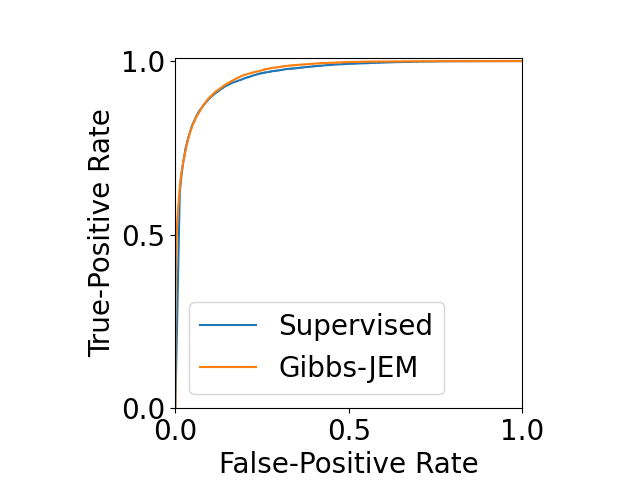}
    \includegraphics[width=.23\textwidth, clip, 
    trim=2.2cm .1cm 2.4cm 1.2cm]
    {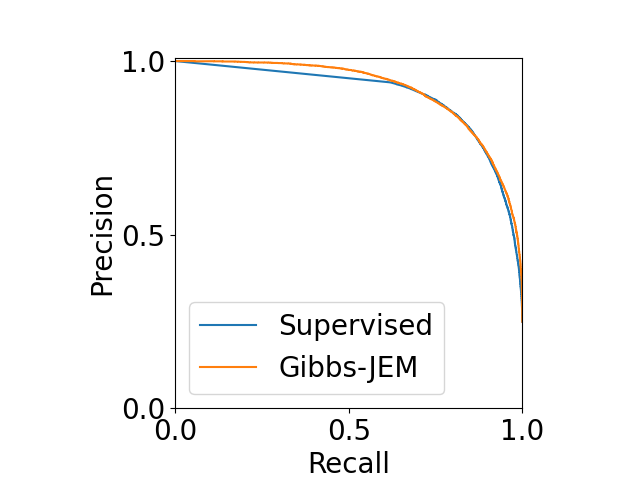}
    \vspace{-1em}
    \caption{Micro-averaged Receiver Operating Characteristic (left) and Precision-Recall (right) curves on UT Zappos50K.}
\end{figure}

\begin{figure}[h]
    \centering
    \includegraphics[width=.23\textwidth, clip, 
    trim=2.2cm .1cm 2.4cm 1.2cm]
    {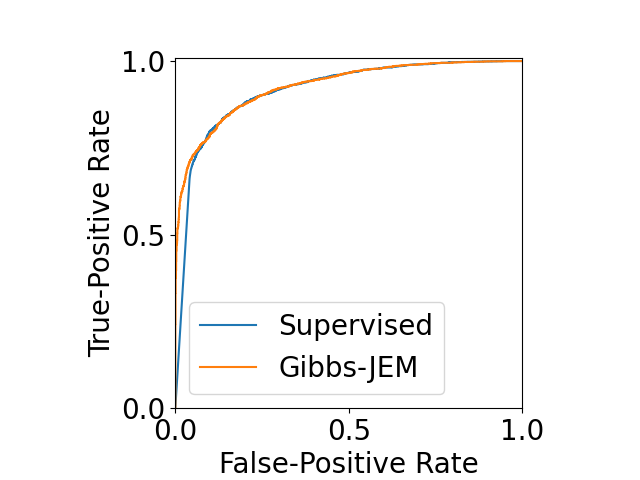}
    \includegraphics[width=.23\textwidth, clip, 
    trim=2.2cm .1cm 2.4cm 1.2cm]
    {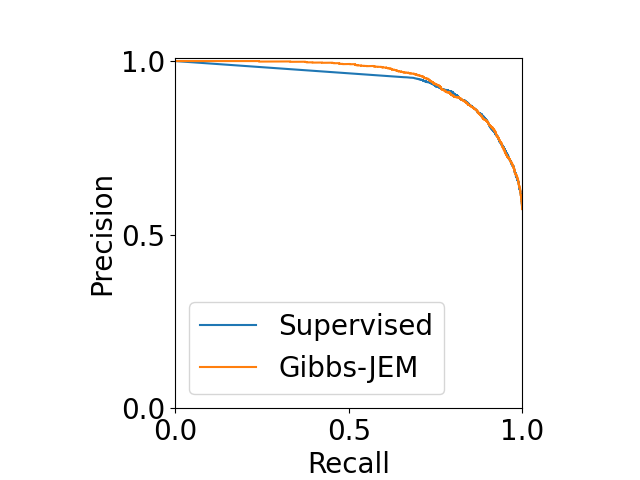}
    \vspace{-1em}
    \caption{Receiver Operating Characteristic (left) and Precision-Recall (right) curves on UT Zappos50K for attribute \enquote{Women's}.}
\end{figure}

\begin{figure}[h]
    \centering
    \includegraphics[width=.23\textwidth, clip, 
    trim=2.2cm .1cm 2.4cm 1.2cm]
    {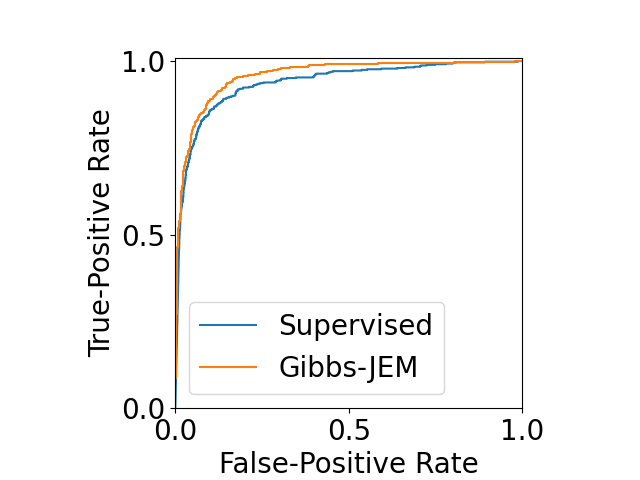}
    \includegraphics[width=.23\textwidth, clip, 
    trim=2.2cm .1cm 2.4cm 1.2cm]
    {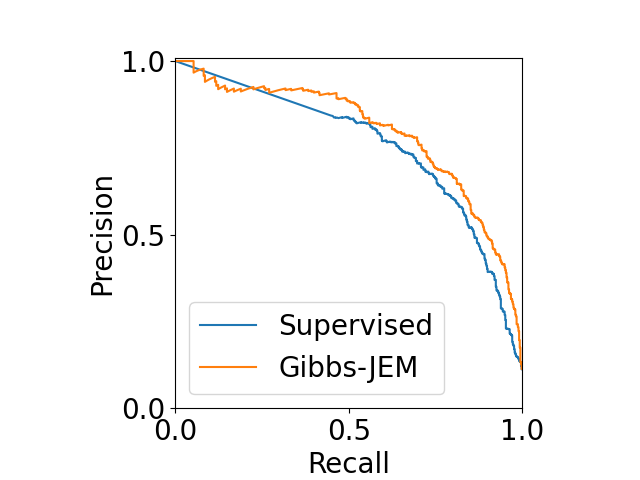}
    \vspace{-1em}
    \caption{Receiver Operating Characteristic (left) and Precision-Recall (right) curves on UT Zappos50K for attribute \enquote{Mesh Material}.}
\end{figure}

\begin{figure}[h]
    \centering
    \includegraphics[width=.23\textwidth, clip, 
    trim=2.2cm .1cm 2.4cm 1.2cm]
    {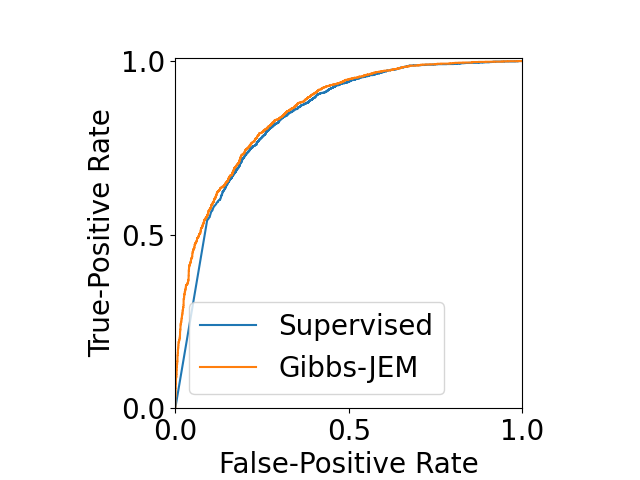}
    \includegraphics[width=.23\textwidth, clip, 
    trim=2.2cm .1cm 2.4cm 1.2cm]
    {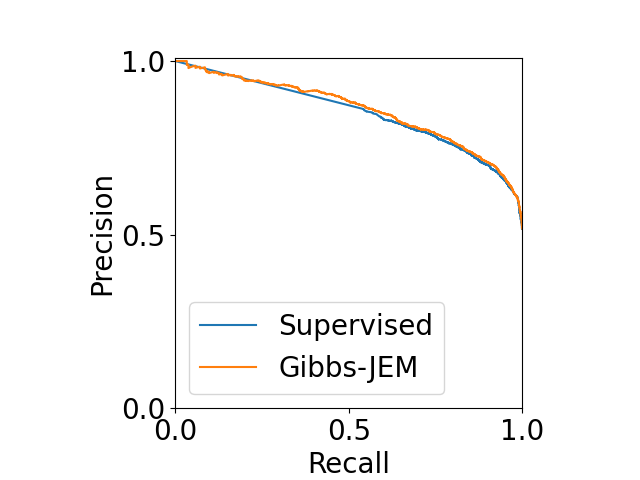}
    \vspace{-1em}
    \caption{Receiver Operating Characteristic (left) and Precision-Recall (right) curves on CelebA for attribute \enquote{Attractive}.}
\end{figure}

\begin{figure}[h]
    \centering
    \includegraphics[width=.23\textwidth, clip, 
    trim=2.2cm .1cm 2.4cm 1.2cm]
    {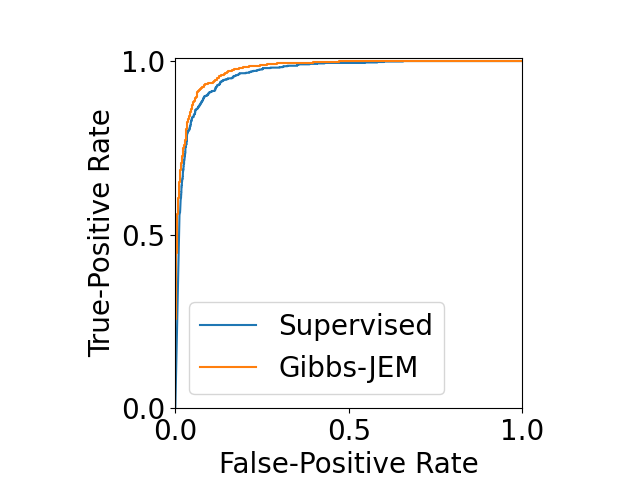}
    \includegraphics[width=.23\textwidth, clip, 
    trim=2.2cm .1cm 2.4cm 1.2cm]
    {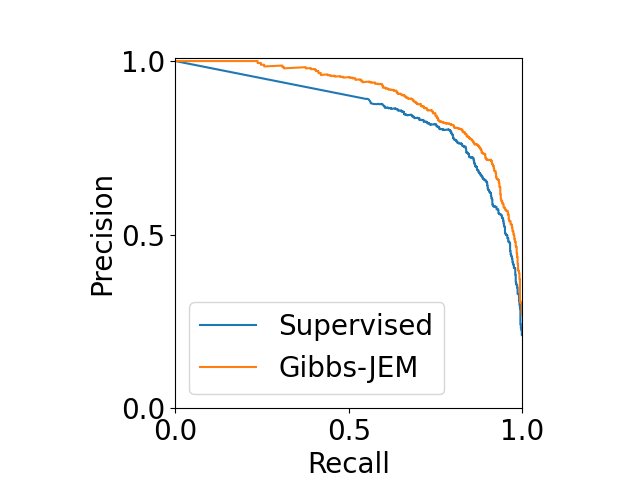}
    \vspace{-1em}
    \caption{Receiver Operating Characteristic (left) and Precision-Recall (right) curves on CelebA for attribute \enquote{Blond Hair}.}
\end{figure}

\section{Additional Samples}
\label{appendix:samples}

\subsection{Data Samples}

\begin{figure}[h]
    \centering
    \includegraphics[width=.45\textwidth, clip]
    {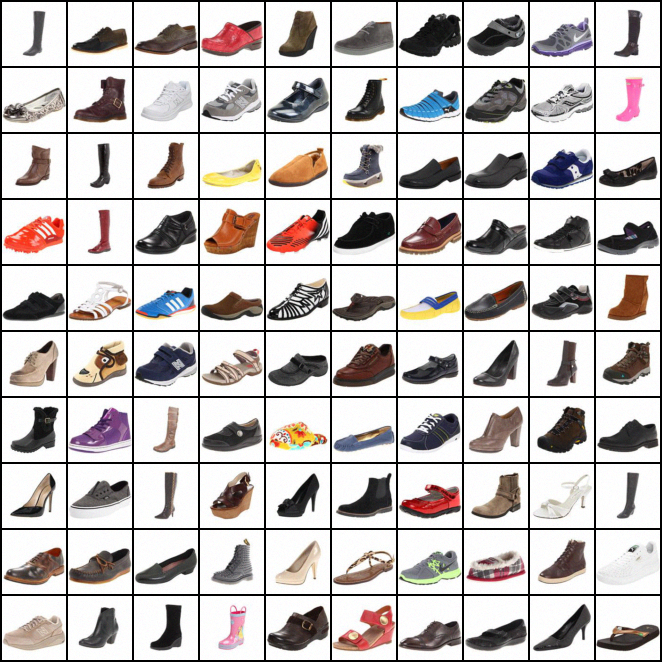}
    \caption{Unconditional data from UT Zappos50K.\label{fig:uncond_data_utzappos}}
\end{figure}

\begin{figure}[h]
    \centering
    \includegraphics[width=.45\textwidth, clip]
    {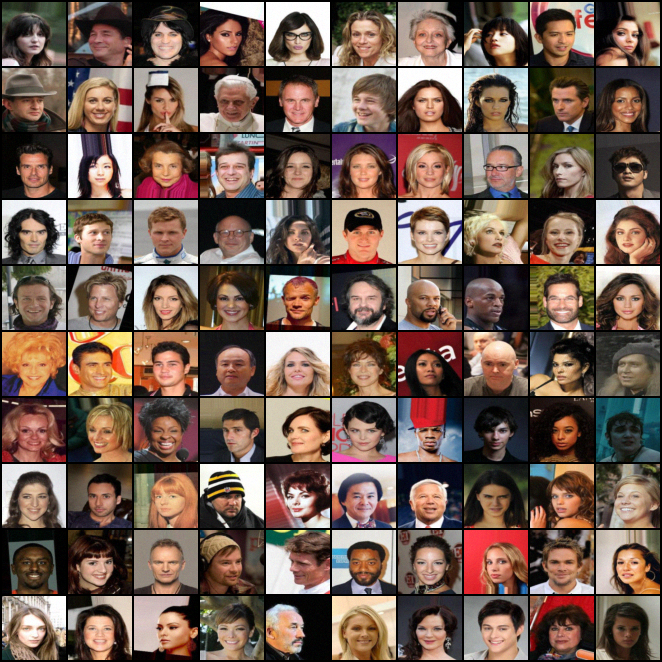}
    \caption{Unconditional data from CelebA.\label{fig:uncond_data_celeba}}
\end{figure}

In Figures \ref{fig:uncond_data_utzappos} and \ref{fig:uncond_data_celeba}, we plot random data samples from both datasets used in our experiments.
 
\subsection{Samples from the Buffer vs. from Noise}

\begin{figure}[h]
    \centering
    \includegraphics[width=.45\textwidth, clip]
    {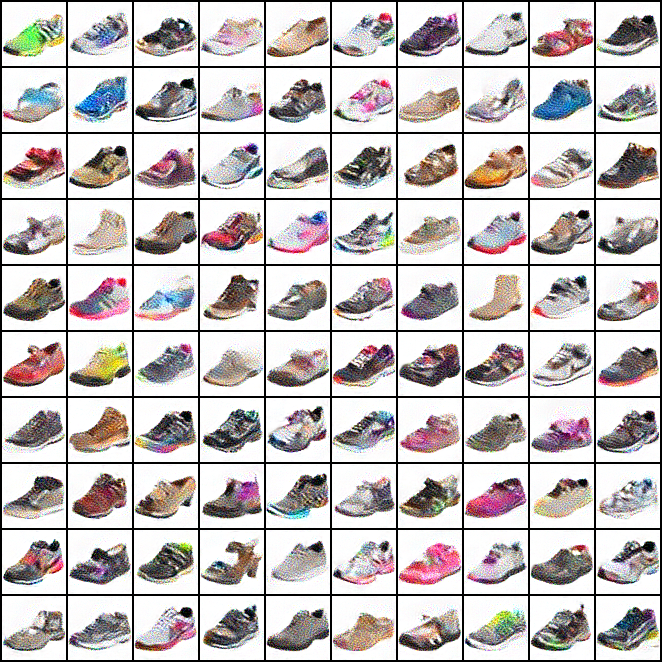}
    \caption{Unconditional samples from noise on UT Zappos50K.\label{fig:uncond_fresh_utzappos}}
\end{figure}

\begin{figure}[h]
    \centering
    \includegraphics[width=.45\textwidth, clip]
    {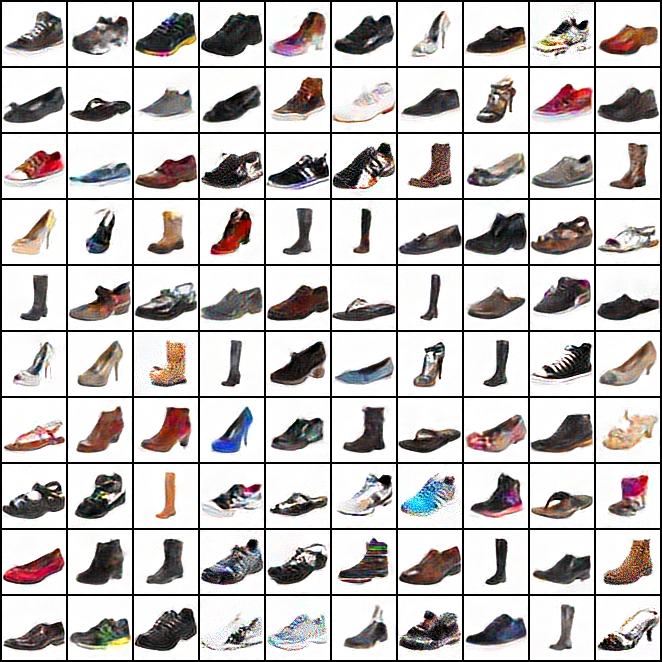}
    \caption{Unconditional samples from buffer on UT Zappos50K.\label{fig:uncond_buffer_utzappos}}
\end{figure}

\begin{figure}[h]
    \centering
    \includegraphics[width=.45\textwidth, clip]
    {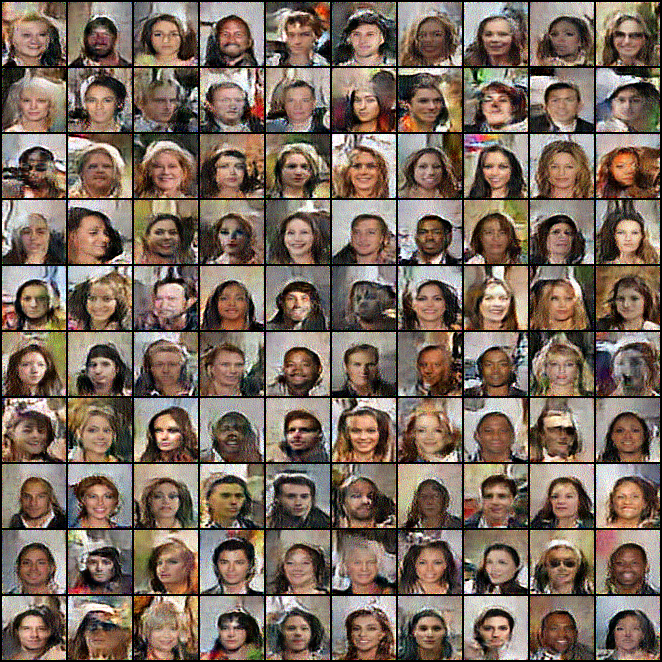}
    \caption{Unconditional samples from noise on CelebA.\label{fig:uncond_fresh_celeba}}
\end{figure}

\begin{figure}[h]
    \centering
    \includegraphics[width=.45\textwidth, clip]
    {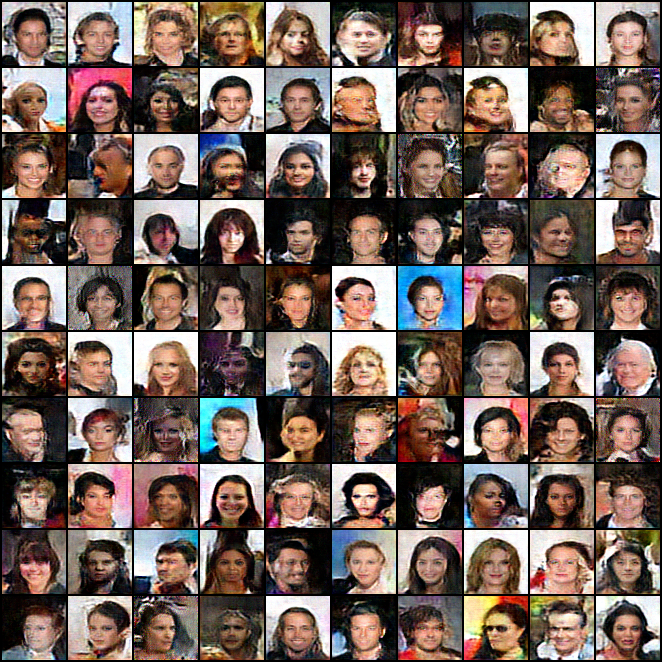}
    \caption{Unconditional samples from buffer on CelebA.\label{fig:uncond_buffer_celeba}}
\end{figure}

\begin{figure}[h]
    \centering
    \includegraphics[width=.45\textwidth, clip]
    {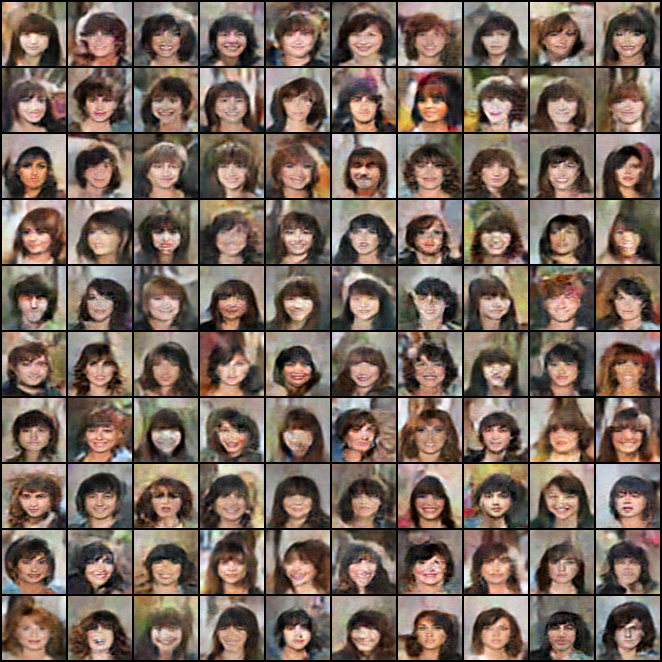}
    \caption{Conditional samples from noise on the attribute \enquote{Bangs} on CelebA.\label{fig:cond_bangs_fresh_celeba}}
\end{figure}

\begin{figure}[h]
    \centering
    \includegraphics[width=.45\textwidth, clip]
    {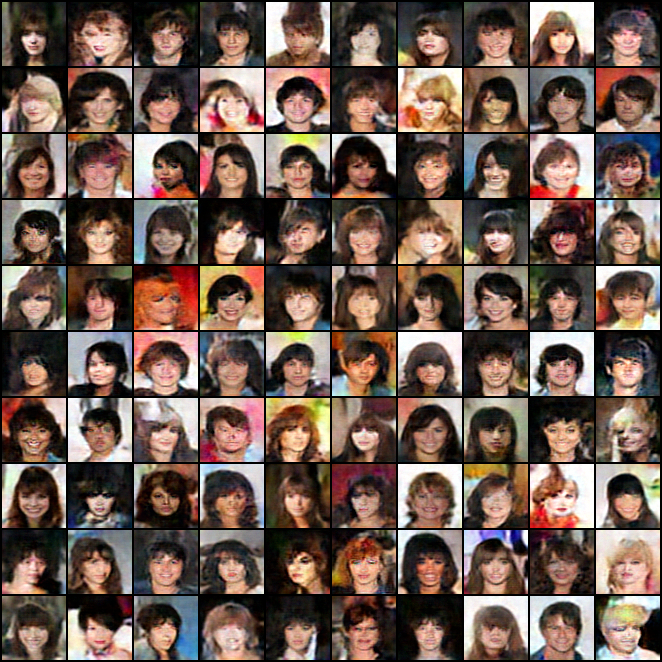}
    \caption{Conditional samples from buffer on the attribute \enquote{Bangs} on CelebA.\label{fig:cond_bangs_buffer_celeba}}
\end{figure}

% \begin{figure}[h]
%     \centering
%     \includegraphics[width=.45\textwidth, clip]
%     {celeba/buffer_vs_noise/samples_custom_12_1_14_0_cond_fresh_26001.png}
%     \caption{Conditional samples from noise on the attribute \enquote{Male} and \enquote{Beard} on CelebA.\label{fig:cond_male_beard_fresh_celeba}}
% \end{figure}

% \begin{figure}[h]
%     \centering
%     \includegraphics[width=.45\textwidth, clip]
%     {celeba/buffer_vs_noise/samples_custom_12_1_14_0_cond_continue_buffer_26001.png}
%     \caption{Conditional samples from buffer on the attribute \enquote{Male} and \enquote{Beard} on CelebA.\label{fig:cond_male_beard_buffer_celeba}}
% \end{figure}

During training with Persistent Contrastive Divergence (PCD), sampling chains are initialized from a replay buffer of chains. At test-time, we have the option of initializing samples from fresh noise, or from samples in the replay buffer used during training.

In Figures \ref{fig:uncond_fresh_utzappos} through \ref{fig:uncond_buffer_celeba}, we plot unconditional samples from our model from fresh noise and the buffer. We find that samples from fresh noise are of slightly lower quality and slightly less diverse than samples from the buffer, especially in the backgrounds of the images. We found that the techniques from \citet{du2020improved} significantly improved the quality and diversity of samples from noise. In Figures \ref{fig:cond_bangs_fresh_celeba}, \ref{fig:cond_bangs_buffer_celeba}, we compare fresh and buffer samples when conditioning on attributes. 
% We observe a similar trend, though notably in Figure \ref{fig:cond_male_beard_fresh_celeba} samples from noise diverge more severely than from the buffer in Figure \ref{fig:cond_male_beard_buffer_celeba}. 
One approach we found useful for mitigating diverged samples was to filter samples using the model's likelihood. In particular, for a desired batch size $N$ of samples, we generate $10N$ samples, then score each sample using the conditional $p_\theta(y_c|x)$ of our model, where $y_c$ are the conditioning attributes. We then take the top $N$ samples with the highest scores.

An important question to consider when initializing chains from the buffer for conditional sampling is whether there are samples of the desired attribute combination are available. We use a buffer of size $10,000$ throughout our experiments here, but CelebA has many more unique attribute combinations than there are samples in the buffer. Thus we cannot always rely upon initializing from the buffer to performing conditional sampling. On the other hand, samples generated from the buffer tend to be more diverse and of higher quality. 
so one question to ask is whether we can initialize unconditionally from the buffer for conditional sampling. That is, we take samples from the buffer without filtering to samples with the conditioned attributes. This way, we can retain the benefits of improved sample quality while also conditioning on attributes not in the buffer. To accomplish this, we would need a model and sampling procedure which effectively mixes between modes with different attribute settings. 

\begin{figure}[h]
    \centering
    \includegraphics[width=.45\textwidth, clip]
    {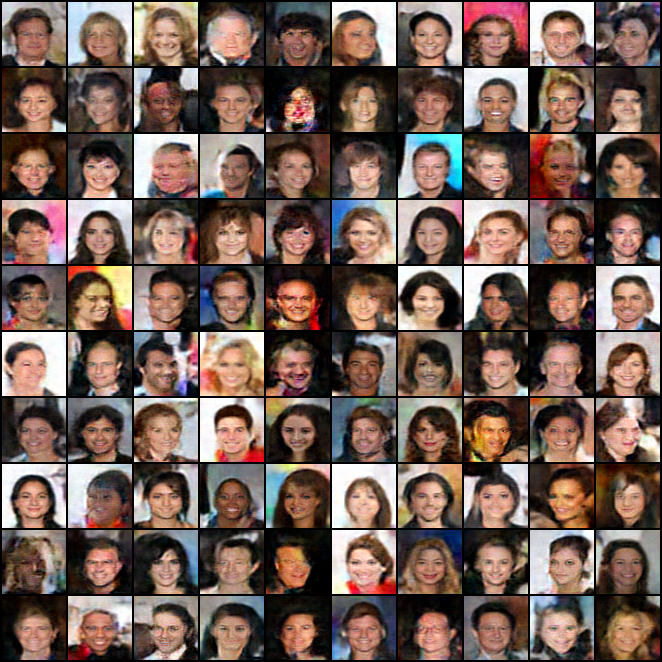}
    \caption{Conditional samples initialized conditionally from buffer on the attribute \enquote{Smiling} on CelebA.\label{fig:cond_smiling_cond_buffer_celeba}}
\end{figure}

\begin{figure}[h]
    \centering
    \includegraphics[width=.45\textwidth, clip]
    {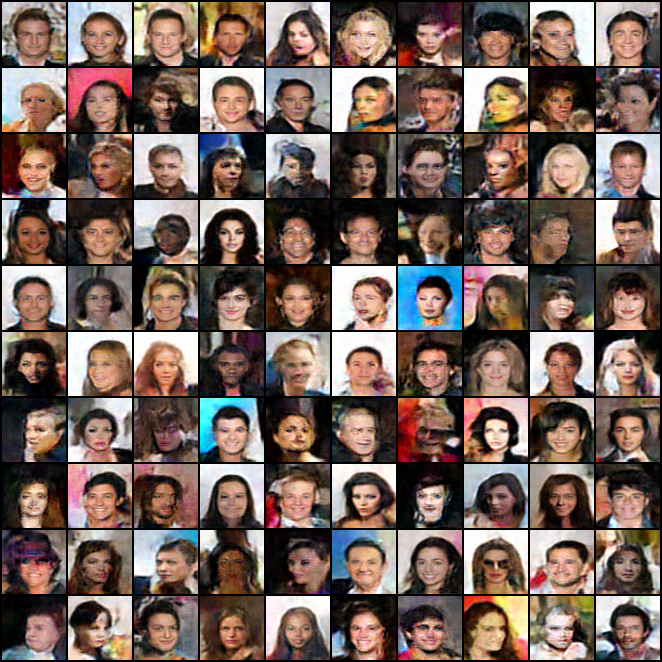}
    \caption{Conditional samples initialized unconditionally from buffer on the attribute \enquote{Smiling} on CelebA.\label{fig:cond_smiling_uncond_buffer_celeba}}
\end{figure}

In Figures \ref{fig:cond_smiling_cond_buffer_celeba}, \ref{fig:cond_smiling_uncond_buffer_celeba}. we examine conditional sampling when sampling conditionally and unconditionally from the buffer, respectively. We find that sampling is unable to consistently to mix well between modes with different attributes.

% \begin{figure}[h]
%     \centering
%     \includegraphics[width=.45\textwidth, clip]
%     {celeba/uncond_buffer/samples_3_1_cond_continue_buffer_26001.png}
%     \caption{Conditional samples initialized conditionally from buffer on the attribute \enquote{Bangs} on CelebA.\label{fig:cond_bangs_cond_buffer_celeba}}
% \end{figure}

% \begin{figure}[h]
%     \centering
%     \includegraphics[width=.45\textwidth, clip]
%     {celeba/uncond_buffer/samples_3_1_cond_continue_buffer_uncond_26001.png}
%     \caption{Conditional samples initialized unconditionally from buffer on the attribute \enquote{Bangs} on CelebA.\label{fig:cond_bangs_uncond_buffer_celeba}}
% \end{figure}

\subsection{Conditional Sampling via Resampling vs. Marginalization}

\begin{figure}[h]
    \centering
    \includegraphics[width=.45\textwidth, clip]
    {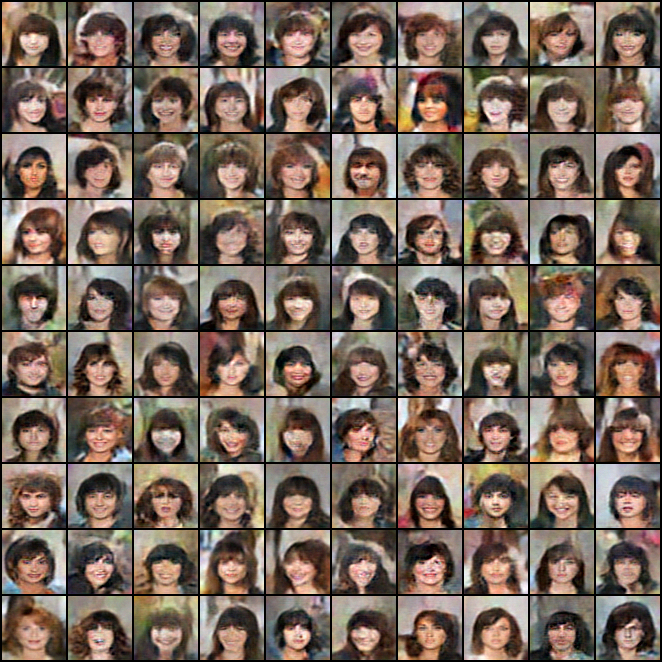}
    \caption{Conditional samples via resampling from noise on the attribute \enquote{Bangs} on CelebA.\label{fig:cond_resample_bangs_fresh_celeba}}
\end{figure}

\begin{figure}[h]
    \centering
    \includegraphics[width=.45\textwidth, clip]
    {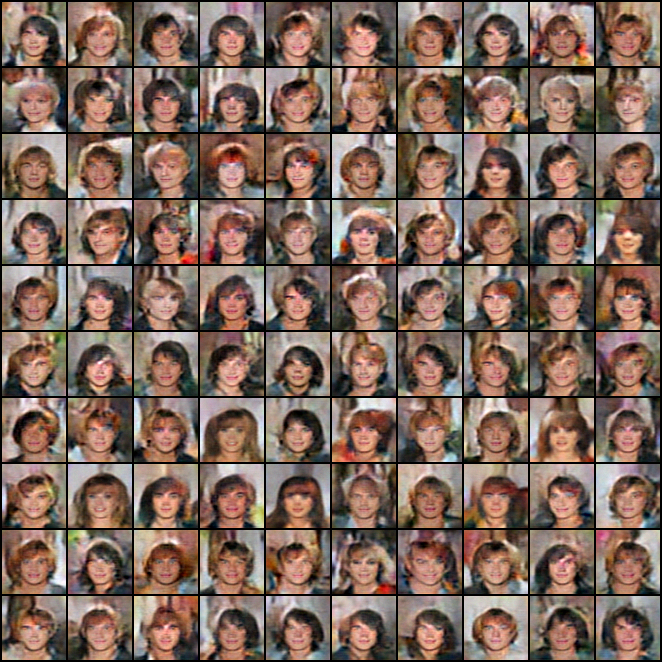}
    \caption{Conditional samples via marginalization from noise on the attribute \enquote{Bangs} on CelebA.\label{fig:cond_marginal_bangs_fresh_celeba}}
\end{figure}

\begin{figure}[h]
    \centering
    \includegraphics[width=.45\textwidth, clip]
    {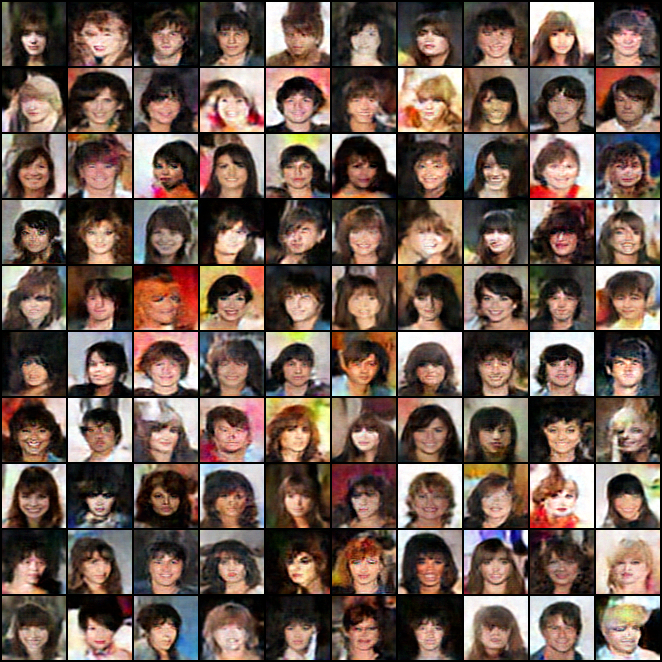}
    \caption{Conditional samples via resampling from buffer on the attribute \enquote{Bangs} on CelebA.\label{fig:cond_resample_bangs_buffer_celeba}}
\end{figure}

\begin{figure}[h]
    \centering
    \includegraphics[width=.45\textwidth, clip]
    {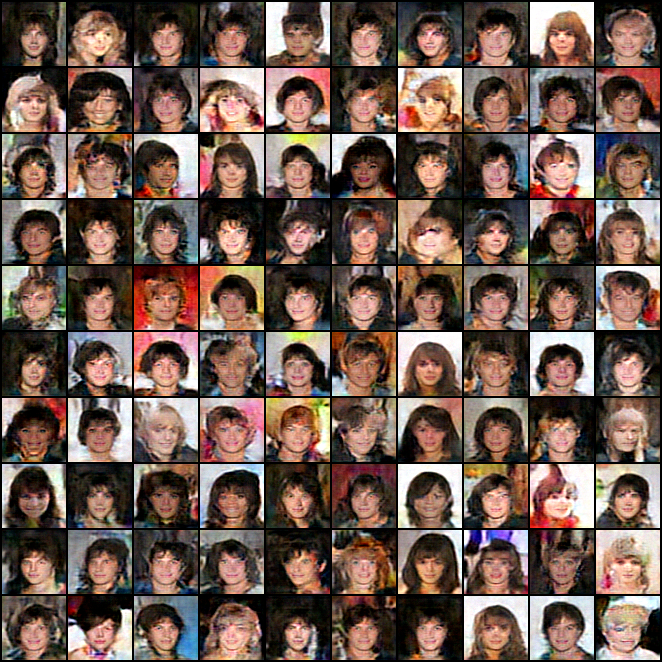}
    \caption{Conditional samples via marginalization from buffer on the attribute \enquote{Bangs} on CelebA.\label{fig:cond_marginal_bangs_buffer_celeba}}
\end{figure}

\begin{figure}[h]
    \centering
    \includegraphics[width=.45\textwidth, clip]
    {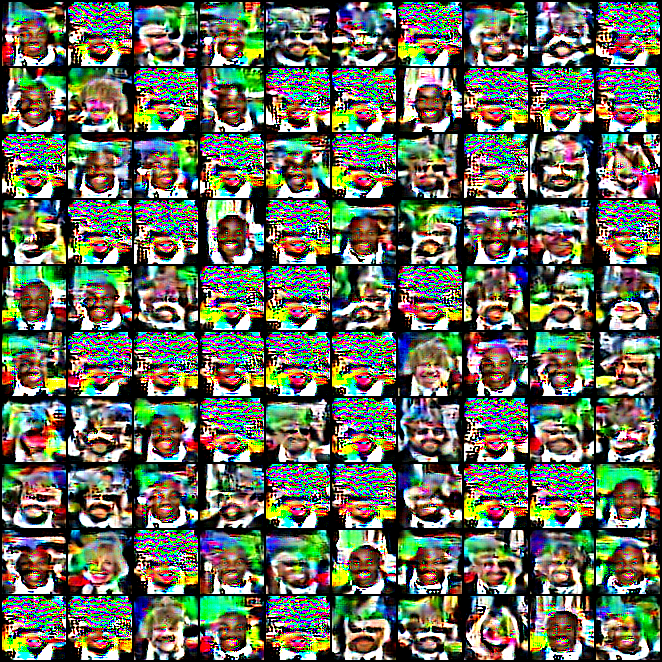}
    \caption{JEM conditional samples via marginalization from noise on the attribute \enquote{Smiling} on CelebA.\label{fig:jem_cond_marginal_smiling_fresh_celeba}}
\end{figure}

As discussed in Appendix \ref{appendix:jem}, we can draw conditional samples either by resampling or marginalizing out the free attributes $y$. In Figures \ref{fig:cond_marginal_bangs_fresh_celeba},
\ref{fig:cond_resample_bangs_fresh_celeba},
\ref{fig:cond_marginal_bangs_buffer_celeba},
\ref{fig:cond_resample_bangs_buffer_celeba}, we compare each sampling method on both fresh noise and resampling. One surprising result is that both sampling methods gives decent sample quality. In particular it is surprising that the bias in MCMC sampling as reported in \citet{nijkamp2020learning} does not cause sampling via marginalization to diverge, despite resampling being the sampling approach used during training. It is apparent however that marginalizing and resampling give qualitatively different samples. We found both sampling methods to work well for our application, and used both methods in different conditioning settings. In contrast, in Figure \ref{fig:jem_cond_marginal_smiling_fresh_celeba}, we find that JEM is unable to generate conditional samples using marginalizing. We observed similar results when using resampling. This confirms our earlier claim that the only reliable method to generate conditional samples is via the factorization $p_\theta(x|y) \propto p_\theta(x)p_\theta(y|x)$. That is, we can generate conditional samples from JEM only by sampling unconditionally from the model, and then classifying the samples using the $p_\theta(y|x)$ model.

% \begin{figure}[h]
%     \centering
%     \includegraphics[width=.45\textwidth, clip]
%     {celeba/resample_vs_marginal/jem/samples_17_1_cond_fresh_28001.png}
%     \caption{JEM conditional samples via resampling from noise on the attribute \enquote{Sampling} on CelebA.\label{fig:jem_cond_resample_smiling_fresh_celeba}}
% \end{figure}

\subsection{How accurate is conditional sampling?}

One sanity check for conditional sampling is to ensure that conditional samples agree with the conditional $p_\theta(y|x)$. That is, if we generate samples conditioned on attributes $y_c$ via $p_\theta(x | y_c)$, how do we know those samples have the attribute $y_c$? The main method we used for determining this was visual inspection of the samples. A slightly better approach would be to use the supervised baseline to classify conditional samples, and measure the accuracy. Here we verify that the conditional samples are internally consistent. That is, we verify that when we generate samples via $p_\theta(x|y_c)$, our model classifies the samples as belong to $y_c$ when using classifier of the same model $p_\theta(y_c|x)$. In particular, when conditioning on pairs of attributes, our model's conditional samples were classified correctly according to the model with an accuracy of $>97\%$ over 100 samples for each pair conditioning. We found this result held across sampling from noise and from the buffer, and using resampling or marginalization for sampling.

\subsection{Synthesizing Novel Attribute Combinations}

One of the main motivations for our work is the ability of our model to synthesize novel attribute combinations. Here we investigate in more detail the performance of our model on this task.

As discussed previously, traditional Factored JEM models can only generate conditional samples by sampling unconditionally and then classifying the resulting samples. This method becomes very inefficient when we wish to sample attribute combinations not seen during training. On the 11 paired attribute combinations we held out during training, all novel attribute combinations occurred in less than $1\%$ of samples. Similarly, we found that the held out attribute combinations occurred in less than $1\%$ of samples in the buffer. While rare, we still have samples of novel attribute combinations in the buffer. We find however that these samples are not useful for initializing sampling chains and lead to poor sampling quality. We suspect this is because the samples from the buffer with the novel attributes have either diverged or are incorrectly classified by the model.

\begin{figure}[h]
    \centering
    \includegraphics[width=.45\textwidth, clip]
    {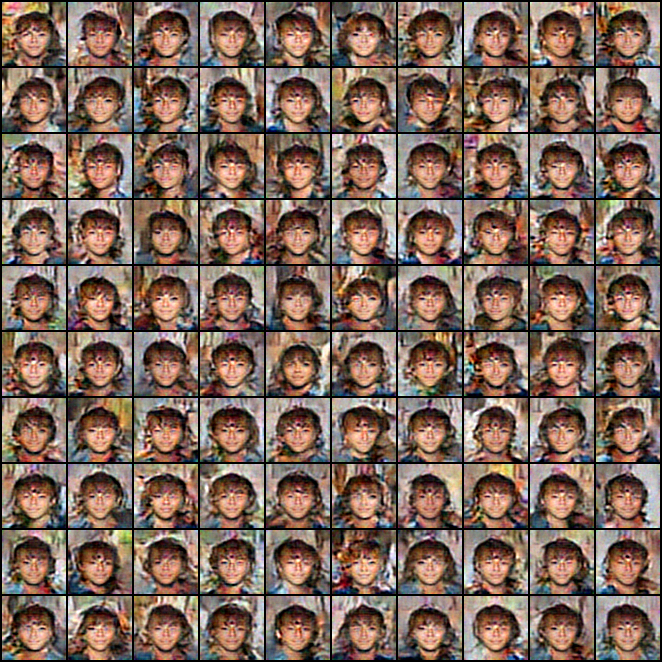}
    \caption{Conditional samples via marginalization from noise on the attribute \enquote{Bangs} on CelebA-novel.\label{fig:cond_marginal_smiling_fresh_celeba_novel}}
\end{figure}

\begin{figure}[h]
    \centering
    \includegraphics[width=.45\textwidth, clip]
    {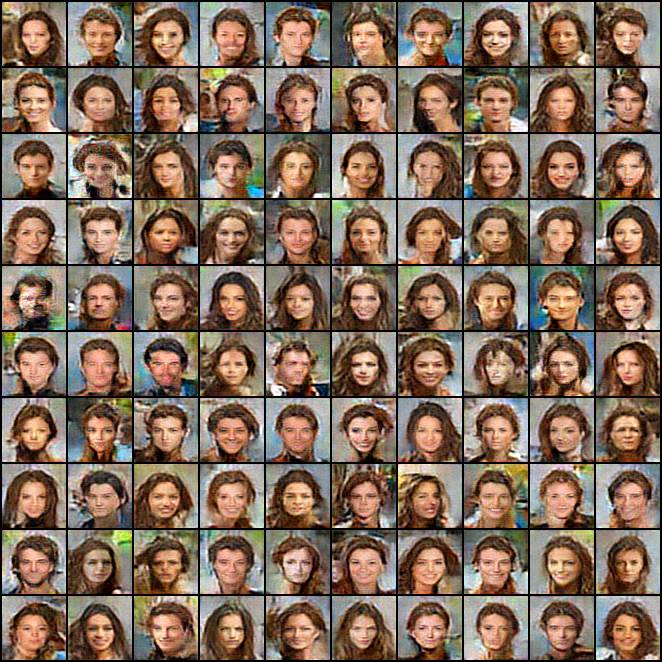}
    \caption{Conditional samples via resampling from noise on the attribute \enquote{Brown Hair} on CelebA-novel.\label{fig:cond_resample_brown_fresh_celeba_novel}}
\end{figure}

\begin{figure}[h]
    \centering
    \includegraphics[width=.45\textwidth, clip]
    {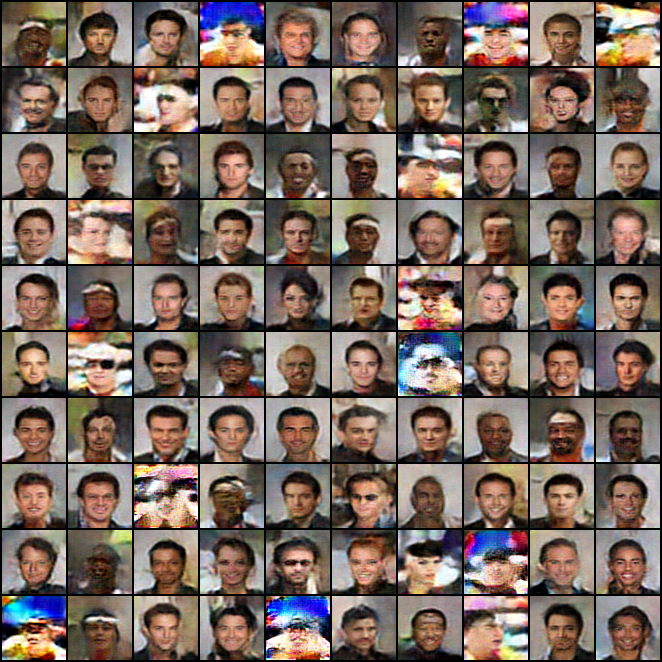}
    \caption{Conditional samples via marginalizing from noise on the attribute \enquote{Male} on CelebA-novel.\label{fig:cond_marginal_male_fresh_celeba_novel}}
\end{figure}

\begin{figure}[h]
    \centering
    \includegraphics[width=.45\textwidth, clip]
    {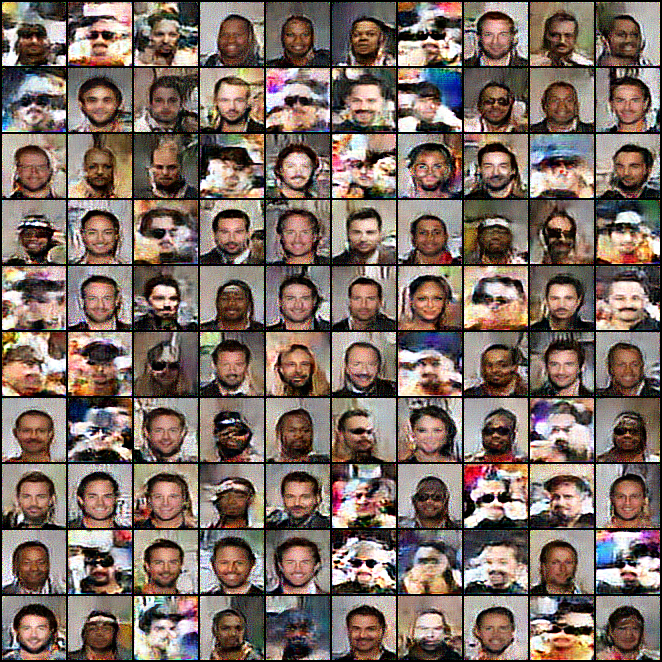}
    \caption{Conditional samples via marginalizing from noise on the attribute \enquote{Beard} on CelebA-novel.\label{fig:cond_marginal_beard_fresh_celeba_novel}}
\end{figure}

\begin{figure}[h]
    \centering
    \includegraphics[width=.45\textwidth, clip]
    {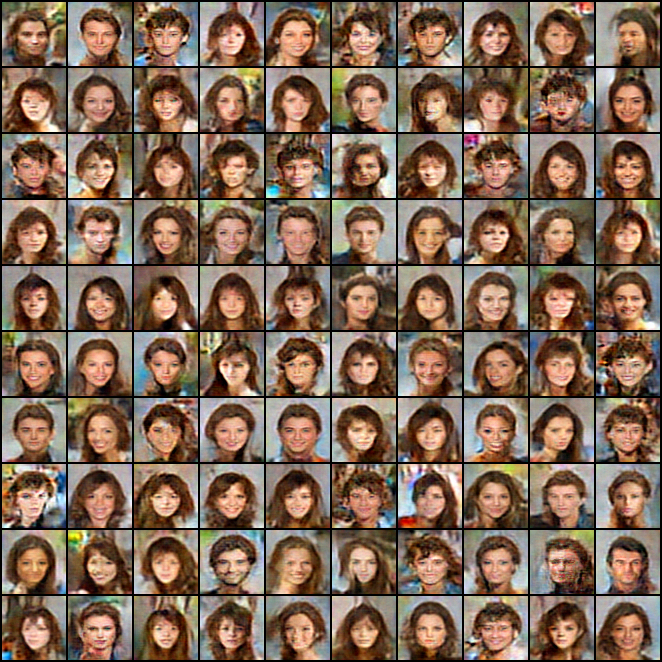}
    \caption{Conditional samples via resampling from noise on the held-out attributes \enquote{Bangs} and \enquote{Brown Hair} on CelebA-novel.\label{fig:cond_resample_bangs_brown_fresh_celeba_novel}}
\end{figure}

\begin{figure}[h]
    \centering
    \includegraphics[width=.45\textwidth, clip]
    {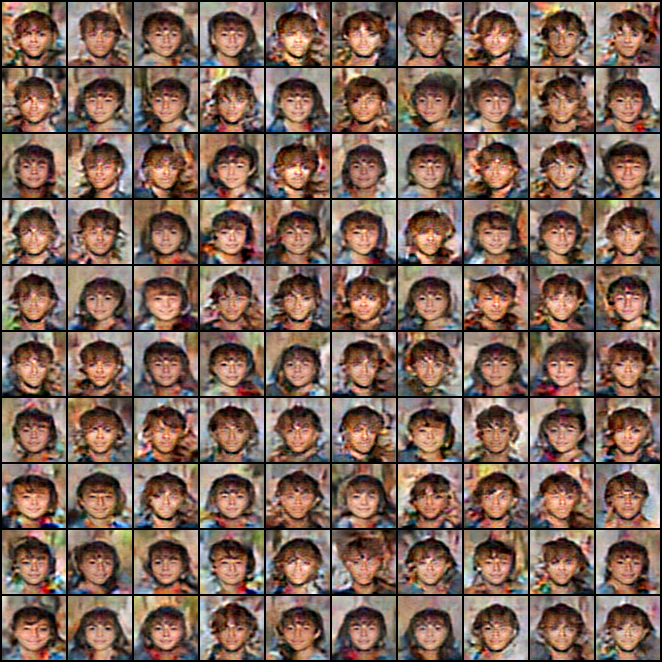}
    \caption{Conditional samples via marginalizing from noise on the held-out attributes \enquote{Bangs} and \enquote{Male} on CelebA-novel.\label{fig:cond_marginal_bangs_male_fresh_celeba_novel}}
\end{figure}

\begin{figure}[h]
    \centering
    \includegraphics[width=.45\textwidth, clip]
    {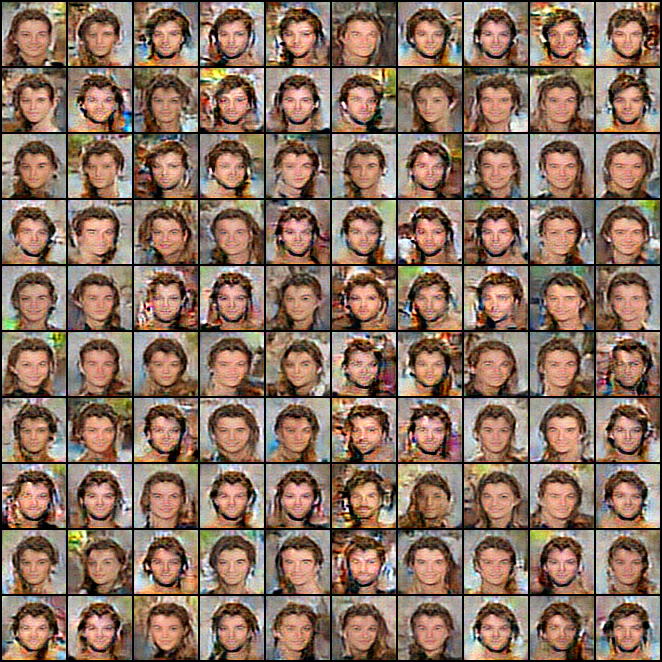}
    \caption{Conditional samples via marginalizing from noise on the held-out attributes \enquote{Beard} and \enquote{Brown Hair} on CelebA-novel.\label{fig:cond_marginal_beard_brown_fresh_celeba_novel}}
\end{figure}

We refer to CelebA with the held-out attribute combinations as Celeba-novel. In Figures \ref{fig:cond_marginal_smiling_fresh_celeba_novel} through \ref{fig:cond_marginal_beard_fresh_celeba_novel} we show samples for the individual attributes whose combinations are held-out for test-time. We found that sample quality differed more significantly here between resampling and marginalization, and chose the option which looked the best for each attribute combination. In Figures \ref{fig:cond_resample_bangs_brown_fresh_celeba_novel} through \ref{fig:cond_marginal_beard_brown_fresh_celeba_novel}, we show samples on novel attribute combinations.

% \begin{figure}[h]
%     \centering
%     \includegraphics[width=.45\textwidth, clip]
%     {celeba/zero shot/samples_custom_3_1_8_1_cond_fresh_marginalize_23001.png}
%     \caption{Conditional samples via marginalizing from noise on the held-out attributes \enquote{Bangs} and \enquote{Brown Hair} on CelebA-novel.\label{fig:cond_marginal_bangs_brown_fresh_celeba_novel}}
% \end{figure}

% \begin{figure}[h]
%     \centering
%     \includegraphics[width=.45\textwidth, clip]
%     {celeba/zero shot/samples_custom_12_1_3_1_cond_fresh_23001.png}
%     \caption{Conditional samples via resampling from noise on the held-out attributes \enquote{Bangs} and \enquote{Male} on CelebA-novel.\label{fig:cond_resample_bangs_male_fresh_celeba_novel}}
% \end{figure}

% \begin{figure}[h]
%     \centering
%     \includegraphics[width=.45\textwidth, clip]
%     {celeba/zero shot/samples_custom_14_0_8_1_cond_fresh_23001.png}
%     \caption{Conditional samples via resampling from noise on the held-out attributes \enquote{Beard} and \enquote{Brown Hair} on CelebA-novel.\label{fig:cond_resample_beard_brown_fresh_celeba_novel}}
% \end{figure}

\end{document}